\documentclass[3p, 10pt, preprint, twocolumn]{elsarticle}

\usepackage{amsmath,amssymb,amsthm,amsfonts}
\usepackage{booktabs}
\usepackage{multirow}
\usepackage{graphicx}
\usepackage{hyperref}
\usepackage{xcolor}
\usepackage{nicefrac}
\usepackage{microtype}
\usepackage{algorithm}
\usepackage{algpseudocode}
\usepackage{tikz}
\usetikzlibrary{arrows.meta,decorations.pathreplacing}

\definecolor{mygold}{RGB}{181,121,44}
\definecolor{myblue}{HTML}{3A5A98}
\definecolor{myred}{RGB}{207, 58, 36}
\definecolor{mygreen}{HTML}{4A8A5C}

\biboptions{sort&compress}

\journal{Neurocomputing}

\newcommand{\F}{\mathcal{F}}

\newcommand{\xx}{\mathbf{x}}
\newcommand{\yy}{\mathbf{y}}
\newcommand{\zz}{\mathbf{z}}

\newtheorem{proposition}{Proposition}

\begin{document}

\begin{frontmatter}

\title{Learning Long-Range Dependencies with Temporal Predictive Coding}

\author[man]{Tom Potter\corref{cor1}}
\ead{thomas.potter@manchester.ac.uk}

\author[man]{Oliver Rhodes}
\ead{oliver.rhodes@manchester.ac.uk}

\cortext[cor1]{Corresponding author}

\address[man]{International Centre for Neuromorphic Systems, The University of Manchester, Manchester, United Kingdom}

\begin{abstract}
Temporal Predictive Coding provides a layer-local, parallelisable mechanism for learning in recurrent systems, making it an attractive candidate for online local learning on neuromorphic and edge hardware. However, its recurrent parameter update captures only local temporal relationships, neglecting the historic influence of parameters along the latent-state trajectory, and therefore struggles to assign credit over longer temporal horizons. This work combines for the first time Temporal Predictive Coding with Real-Time Recurrent Learning (tPC-RTRL), incorporating an online influence matrix that tracks this historic effect whilst preserving the spatial and temporal locality properties valued by neuromorphic implementations. Under explicit assumptions, we prove that tPC-RTRL recovers the gradients of backpropagation-through-time exactly. Empirically, a near-equivalence holds across several tasks of varying scale and complexity, including byte-level language modelling on WikiText-103 (tPC-RTRL vs.\ BPTT: $1.865$ vs.\ $1.864$ validation BPC), English--French translation on a CCMatrix subset ($20.23$ vs.\ $20.29$ BLEU), and a realistic nanodrone system-identification benchmark ($0.506$\,m vs.\ $0.505$\,m mean position error). Finally, we show that the iterative inference mechanism used during training can be reused at deployment time to incorporate intermittent state observations, halving final-position error relative to open-loop rollout on the nanodrone task ($0.402$\,m vs.\ $0.805$\,m) and suggesting a path towards unifying learning and filtering within the same computational framework.
\end{abstract}


\end{frontmatter}

\section{Introduction}
Backpropagation of errors (BP) \cite{rumelhart1986learning} has long served as the dominant algorithm for training artificial neural networks, powering breakthroughs in computer vision \cite{10.5555/2999134.2999257}, natural language processing \cite{DBLP:journals/corr/abs-1810-04805}, and reinforcement learning \cite{mnih2013playingatarideepreinforcement}. However, as models continue to grow to billions of parameters, training them is becoming prohibitively expensive, both computationally and energetically. This has led to growing concerns about the environmental impact and long-term sustainability of the field \cite{DBLP:journals/corr/abs-1907-10597}, and has prompted the search for learning algorithms with lower resource demands.

Motivated by the efficiency of biological intelligence, neuroscience-inspired alternatives to BP have started to gain traction. The brain operates under strict resource constraints, reasoning and learning with a fraction of the energy of its artificial counterparts. Inspired by this, several recent works have explored the potential of biologically inspired learning algorithms as a pathway towards more efficient learning systems \cite{xie2003equivalence, scellier2017equilibriumpropagationbridginggap, song_inferring_2024}, many of which are attractive from a hardware perspective due to their compatibility with the design principles of neuromorphic and edge computing platforms \cite{davies2018loihi}.

Predictive Coding (PC) \cite{Rao1999Predictive} is one such promising alternative. With roots in theories of cortical function \cite{Mumford1991,Mumford1992}, information theory \cite{Atick2011,Attneave1954}, and ties to variational inference \cite{Friston2003,Friston2005,Friston2008}, PC presents a method of credit assignment consisting of local, parallelisable operations, making it well suited to hardware where inter-core communication dominates energy cost, on-chip memory is limited, and strict global synchronisation is expensive \cite{Orchard2021Loihi2}. Recent work has shown that PC can closely approximate BP-derived updates \cite{Whittington2017AnAO,DBLP:journals/corr/abs-2006-04182} and, under explicit conditions, match them exactly \cite{NEURIPS2020_fec87a37,salvatori2023reversedifferentiationpredictivecoding}. When these exact-equivalence constraints are relaxed, PC has been associated with further desirable properties including improved sample efficiency, robustness, and advantages in online settings \cite{song_inferring_2024}, with recent analyses suggesting that PC-derived updates can capture aspects of higher-order optimisation \cite{mali2024tightstabilityconvergencerobustness}.

Despite this progress, PC approaches remain less developed in recurrent and temporally extended settings. Although several works have explored recurrent PC formulations \cite{Millidge2023.05.15.540906,tang2023sequentialmemorytemporalpredictive,10.1371/journal.pcbi.1011801, 8963851}, they have typically only been evaluated on relatively small-scale tasks \cite{8963851, 10.1371/journal.pcbi.1011801} and often struggle with long-range temporal credit assignment \cite{tang2023sequentialmemorytemporalpredictive,Millidge2023.05.15.540906}.

To address these limitations, this paper explores Temporal Predictive Coding with Real-Time Recurrent Learning (tPC-RTRL): a PC algorithm for recurrent neural networks that combines the local predictive-error computations of PC, with an RTRL-based online recurrent credit-assignment mechanism. We begin by highlighting a specific, structural limitation of standard Temporal Predictive Coding (tPC), identifying that its parameter update rule neglects the historic influence of the recurrent parameters throughout the latent-state trajectory (Section \ref{sec:tpc_background}). We then introduce tPC-RTRL as a natural solution to this problem aiming to improve the model's ability to learn temporal relationships (Section \ref{sec:tpcrtrl_derivation}). Next, we show that, under explicit assumptions, this algorithm recovers BPTT gradients exactly (Section \ref{sec:bptt_relation}), supporting the idea that tPC-RTRL has the ability to recreate the strong performance of BPTT on a range of tasks. Finally, we provide empirical evidence for this claim, demonstrating that tPC-RTRL matches the performance of BPTT on a range of sequence learning tasks (Section \ref{sec:results}).

In summary, this work makes the following contributions:
\begin{itemize}
    \item Introduces Temporal Predictive Coding with Real-Time Recurrent Learning (tPC-RTRL).
    \item Shows that, under explicit assumptions, tPC-RTRL recovers the gradients computed by backpropagation-through-time (BPTT), providing a theoretical validation that BPTT-derived updates can be recovered in a local, parallelisable manner.
    \item Demonstrates the effectiveness of tPC-RTRL on both synthetic and real-world sequence-learning tasks, including language modelling on WikiText-103 \cite{merity2016pointersentinelmixturemodels} and translation on an English--French subset of CCMatrix \cite{schwenk2020ccmatrixminingbillionshighquality}.
    \item Demonstrates that the same iterative inference mechanism used during tPC-RTRL training can be reused for test-time state correction on a real-world nanodrone benchmark \cite{Busetto_2026}.
\end{itemize}

\section{Background}
This section provides background to support the methods developed in Section~\ref{sec:method}. We begin by motivating the search for alternatives to backpropagation (BP) through the lens of neuromorphic hardware (Section~\ref{sec:neuromorphic_motivation}). We then review the Predictive Coding (PC) framework (Section~\ref{sec:pc_background}) underlying our approach, with a particular focus on Temporal Predictive Coding (tPC, Section~\ref{sec:tpc_background}). Finally, we introduce Real-Time Recurrent Learning (RTRL, Section~\ref{sec:rtrl}), the recurrent credit-assignment mechanism we later use to extend tPC to longer-range temporal learning.

\subsection{Neuromorphic hardware and on-chip learning}
\label{sec:neuromorphic_motivation}

Neuromorphic hardware offers a promising alternative to conventional accelerators for applications requiring low-power, low-latency intelligence, operating continually at the edge. Rather than concentrating computation in a small number of large processing units, neuromorphic systems distribute state, memory, and computation across many small cores. Each core typically maintains local neuron and synapse state and exchanges information primarily with synaptically connected units. The low-power potential of this computing paradigm is attractive for several applications, for example in always-on acoustic or visual sensing, wearable monitoring, autonomous robotics, drones, and industrial monitoring. In such settings, transmitting raw data to a remote accelerator or repeatedly retraining a model offline may be undesirable because of energy, latency, bandwidth, or privacy constraints \cite{Orchard2021Loihi2,Gonzalez2024SpiNNaker2,Akopyan2015TrueNorth}.

For inference, neural networks can often be mapped relatively directly onto such substrates. Neural state can be maintained locally, and activations are communicated along the same synaptic connections that define the model. However, training these models on-chip is substantially more difficult. BP, the algorithm used to train modern neural networks, consists of a forward computation followed by a distinct reverse-mode computation. The forward pass computes neural activations, whereas the backward pass must propagate error derivatives in the opposite direction through the network, combine them with the local activation derivatives, or stored neuron state, and use the resulting signals to update each parameter. Conventional accelerators are designed to support this process. Intermediate activations can be stored in large global memories, and the same dense linear-algebra units used during the forward pass can be reused during a globally coordinated, layer-by-layer reverse traversal of the model's compute graph. In contrast, neuromorphic substrates are primarily designed to update local neural state, in response to incoming synaptic transmissions. For the system to support BP, it must additionally represent and route error signals, retain or reconstruct the activities and derivatives needed for updates, and schedule this process using an explicit coordination or scheduling mechanism.

Much of the on-chip neuromorphic-learning literature reflects these problems. A common approach is to train a network off-chip using BP, then deploy the learnt parameters to neuromorphic hardware for inference. Hardware-in-the-loop methods take a partial step towards adaptation on the device, running the network on the neuromorphic substrate, but reading state from the chip to compute parameter updates externally using conventional methods \cite{schmitt2017neuromorphic}. Other approaches avoid BP altogether, using local plasticity rules such as spike-timing-dependent plasticity, reward-modulated plasticity, or eligibility-trace-based algorithms such as e-prop \cite{wunderlich2019demonstrating,rostami2022eprop,bellec2020solution}. These methods support on-chip and online adaptation using signals that can be computed and routed locally, but make different trade-offs between learning performance, and hardware compatibility.

Recent work by \citet{renner2024backpropagation} demonstrates that exact BP can nevertheless be implemented fully on-chip on Intel's Loihi processor \cite{davies2018loihi}. However, this result does not just map a conventional feedforward network onto the hardware. The implementation augments the feedforward pathway with multiple additional components, including a dedicated error-propagation pathway, relay populations that preserve activity for subsequent updates, and a control circuit that coordinates the successive stages of forward propagation, error computation, backward propagation, and weight updates. This is an important step towards on-chip BP, but it also highlights the accommodations that must be made to implement BP on this type of hardware, and the overheads that are incurred in doing so. The challenge becomes more pronounced in recurrent systems. BPTT accounts not only for how an error propagates across layers, but also for how parameters influenced the current state through the entire recurrent state trajectory. It does this by storing intermediate activations to accommodate unrolling the recurrent network through time and traversing the unrolled computation in reverse. Training recurrent networks on-chip requires new ways of thinking about temporal credit assignment that avoid the need to store and traverse the full latent-state trajectory \cite{bellec2020solution}.

Predictive Coding (PC) offers a potential algorithm which is more compatible with the design principles of neuromorphic hardware. Rather than separating inference from a distinct reverse-mode phase, PC represents prediction errors within the network and iteratively updates latent states to minimise a shared free energy. The local dynamics that minimise this free energy also disperse prediction errors throughout the network. Parameter updates can then be computed from locally available activities and prediction errors. Therefore, the potential advantage of PC is that credit assignment and inference are embedded in the same dynamical process, rather than requiring a separately scheduled global backward computation. Furthermore, although PC inference is described as gradient descent on a free energy objective, it can be implemented through local synaptic interactions, aligning much more closely with the preferences of neuromorphic hardware than the global, reverse-mode traversal of BP. This perspective has led to a growing interest in PC as a low-resource alternative to BP, and motivates PC for on-chip learning. 

\subsection{Predictive Coding}
Predictive Coding (PC) originated as a theory of hierarchical cortical processing, in which each layer predicts the activity of the layer below and only the residual prediction error is propagated forward \citep{Rao1999Predictive,Mumford1991,Mumford1992}. Over time, this idea has been connected to information-theoretic accounts of efficient coding \cite{Atick2011,Attneave1954}, variational free energy formulations of perception and learning \cite{Friston2003,Friston2005,Friston2008}, and more recently to machine-learning views of local credit assignment \cite{Whittington2017AnAO,NEURIPS2020_fec87a37}. From this perspective, PC provides not only a descriptive theory of neural computation, but also a practical algorithmic framework in which inference and learning are both driven by the minimisation of local prediction errors.


\label{sec:pc_background}
\paragraph{PC as a Machine Learning Algorithm}
When viewed as a machine-learning algorithm, PC performs approximate
Bayesian inference in a hierarchical generative model by minimising
a variational free energy $\F$ \cite{Friston2003,DBLP:journals/corr/abs-2107-12979}. Under Gaussian conditionals and a
mean-field approximation in which each variational factor is sharply
peaked around its current state estimate, $\F$ reduces to a sum of
local squared prediction errors,
\begin{equation}
\F
=
\frac{1}{2}
\sum_{l=0}^{L-1}
\epsilon_l^\top \epsilon_l,
\qquad
\epsilon_l = \xx_l - g_l(\xx_{l+1}; 
\theta_l),
\end{equation}
where $g_l$ denotes the downward prediction function from layer $l+1$ to layer
$l$ with parameters $\theta_l$, and $\xx_0 \equiv \yy$ is the observed
target. PC is typically implemented via an Expectation-Maximisation (EM)-style procedure
\cite{Whittington2017AnAO}. An E-step iteratively updates the latent
activities $\xx_{1:L-1}$ to reduce $\F$ with parameters held fixed,
followed by an M-step that updates parameters by gradient descent on
the same objective with activities held fixed. The resulting
inference and learning rules are layer-local, with each update
depending only on the prediction error at its own layer and the
layer directly below. A full derivation, including the variational
free energy setup and the conditions under which the Hebbian-like
parameter update arises, is given in \ref{app:pc_derivation}.

A growing body of work has shown that PC can closely approximate BP in feedforward networks, recover exact BP gradients under suitable assumptions, and extend beyond layered architectures to more general computation graphs \cite{Whittington2017AnAO,DBLP:journals/corr/abs-2006-04182,NEURIPS2020_fec87a37,salvatori2023reversedifferentiationpredictivecoding,salvatori2022learningarbitrarygraphtopologies}. This has shifted PC from being viewed purely as a neuroscientific theory of cortical function towards a promising framework for local credit assignment. At the same time, recent analyses suggest that when exact BP-equivalence constraints are relaxed, PC exhibits desirable optimisation behaviour, including robustness in online settings and the ability to incorporate higher-order curvature information \cite{song_inferring_2024,mali2024tightstabilityconvergencerobustness}.

Recent work has shown
that practical performance can depend strongly on implementation
choices \cite{pinchetti2024benchmarkingpredictivecodingnetworks}.
A growing line of work has begun to study the optimisation
and architectural details required to train PC networks reliably,
with the aim of developing a repertoire of best practices analogous
to those that have contributed to the success of BP. This includes
work on inference and parameter optimisers \cite{alonso2024understanding}
as well as architectural choices such as activation functions
\cite{frieder2024bad}, and modifications to the inference and learning rules to improve training in deeper networks \cite{qi2025training}.


\subsection{Temporal Predictive Coding}
\label{sec:tpc_background}
\paragraph{Recurrent Predictive Coding}
Several lines of work have extended PC to temporally structured data. One family introduces temporal structure through generalised coordinates of motion, in which hidden states are augmented with higher-order temporal derivatives \cite{Friston2008, DBLP:journals/corr/abs-2107-12979}. These approaches provide a principled account of temporal inference and generalised filtering, but can increase modelling and numerical complexity as progressively higher-order derivatives are retained, and may be sensitive to discretisation error and noisy observations. More recently, dynamic PC models have explored hierarchical spatiotemporal prediction in which higher-level latent variables modulate lower-level transition dynamics \cite{10.1371/journal.pcbi.1011801}. These works demonstrate that PC can support rich temporal structure, but they often rely on specialised architectural mechanisms to parameterise dynamics, making it less clear how temporal credit assignment should be performed in standard recurrent networks.

\paragraph{Temporal Predictive Coding (tPC)}
Temporal Predictive Coding extends PC to sequential data by formulating sequence processing as inference in a recurrent system \cite{Millidge2023.05.15.540906, tang2023sequentialmemorytemporalpredictive}. In this view, the goal is not only to predict future observations, but to infer the latent state of the system as observations arrive over time. \citet{Millidge2023.05.15.540906} formulate this problem using a hidden Markov model with control inputs, in which a latent state evolves according to recurrent dynamics and generates noisy observations. An important feature of this formulation is that it connects PC to classical state estimation. In the linear-Gaussian case, the resulting iterative inference dynamics recover hidden-state tracking behaviour closely related to Kalman filtering, but without explicitly propagating posterior covariance estimates \cite{Millidge2023.05.15.540906}. This makes tPC especially appealing in partially observed settings, where sequential prediction and online state inference must be performed simultaneously within a single recurrent dynamical system. Beyond this, \citet{tang2023sequentialmemorytemporalpredictive} study tPC for sequential memory, showing that the approach can memorise and retrieve sequential inputs accurately with a biologically motivated implementation.

Next, we introduce the tPC framework \cite{Millidge2023.05.15.540906} which serves as the starting point of our method. Consider a tPC network, where the input at timestep $t$ is clamped as $\xx_L^{(t)}$, the target is denoted $\yy^{(t)}$, and a latent state $\xx^{(t)}$ must be inferred. The algorithm itself makes minimal specifications about the structure of the network; the key conceptual move is from a static PC network with no temporal relationships, to one in which previous latent states provide some form of prior over inference at the current timestep. For clarity, we first describe the single-latent-state case; the multilayer extension follows directly and is given in \cite{Millidge2023.05.15.540906}.

The temporal generative model factorises as
\begin{align}
p(&\yy^{(t)}, \xx^{(t)} \mid \xx_L^{(t)}, \hat{\xx}^{(t-1)}; \theta)
=\notag\\
&p(\yy^{(t)} \mid \xx^{(t)}; \theta_{\mathrm{out}})\,
p(\xx^{(t)} \mid \xx_L^{(t)}, \hat{\xx}^{(t-1)}; \theta_{\mathrm{rec}}),
\end{align}
where $\hat{\xx}^{(t-1)}$ denotes the converged latent-state estimate from the previous timestep. Thus, inference at time $t$ must balance finding a latent state that explains the current target whilst remaining consistent with a temporal prediction based on the previous state and current input. Assuming Gaussian conditionals and the sharply-peaked variational approximation used in the static case (\ref{app:pc_derivation}), the instantaneous free energy reduces to
\begin{equation}
\F^{(t)}
=
\tfrac{1}{2}\|\epsilon_x^{(t)}\|^2
+
\tfrac{1}{2}\|\epsilon_y^{(t)}\|^2,
\label{eq:tpc_free_energy}
\end{equation}
where
\begin{align}
&\epsilon_x^{(t)} = \xx^{(t)} - f(\xx_L^{(t)}, \hat{\xx}^{(t-1)}; \theta_{\mathrm{rec}}),\\
&\epsilon_y^{(t)} = \yy^{(t)} - g(\xx^{(t)}; \theta_{\mathrm{out}}).
\end{align}
We refer to the functions $f$ and $g$ as the predictive prior and decoder, respectively. These functions generate predictions of neighbouring variables, which we may represent with 
\begin{align}
\mu_x^{(t)} &= f(\xx_L^{(t)}, \hat{\xx}^{(t-1)}; \theta_{\mathrm{rec}}),\\
\mu_y^{(t)} &= g(\xx^{(t)}; \theta_{\mathrm{out}}).
\end{align}
This decoder function is sometimes called the readout, as it maps the recurrent latent state to the output space. It could also be a multilayer network, in which case multiple intermediate latent states may be inferred as in a static PC network.

As in Section \ref{sec:pc_background}, tPC is implemented by alternating iterative inference (E-step) and parameter updates (M-step) at each timestep. The resulting update rules preserve the locality of standard PC, with the addition of a predictive prior through the previous inferred state, given by $f$.

\paragraph{A structural limitation of standard tPC}
This formulation has an important consequence for the recurrent parameters. The instantaneous free energy $\F^{(t)}$ depends on $\theta_{\mathrm{rec}}$ only through the current transition prediction $\mu_x^{(t)}$. The previous inferred state $\hat{\xx}^{(t-1)}$ is treated as a fixed variable, rather than as a function of $\theta_{\mathrm{rec}}$. The recurrent parameter update therefore takes the form
\begin{equation}
\Delta \theta_{\mathrm{rec}}
=
-\eta \frac{\partial \F^{(t)}}{\partial \theta_{\mathrm{rec}}^{(t)}}
=
-\eta
\frac{\partial \F^{(t)}}{\partial \mu_x^{(t)}}
\frac{\partial \mu_x^{(t)}}{\partial \theta_{\mathrm{rec}}^{(t)}},
\label{eq:tpc_immediate_update}
\end{equation}
where $\theta_{\mathrm{rec}}^{(t)}$ denotes the application of the recurrent parameters at time $t$ and $\mu_x^{(t)}$ denotes the recurrent prediction given by the transition prior. This update captures only the immediate influence of $\theta_{\mathrm{rec}}$ on the current recurrent prediction. For parameters that are not reused through the latent-state trajectory, such as $\theta_\mathrm{out}$, this is sufficient. For recurrent parameters, however, it neglects their historic influence through earlier propagated states, and therefore cannot reliably propagate recurrent parameter credit across more than a single timestep. In this respect, standard tPC has the same temporal credit-assignment structure as one-step truncated BPTT. Section~\ref{sec:copy} gives an empirical demonstration of this equivalence. This missing temporal credit-assignment term is what motivates the RTRL-based approach developed in Section~\ref{sec:method}.

\begin{figure*}[t]
    \centering
    \includegraphics[width=\textwidth]{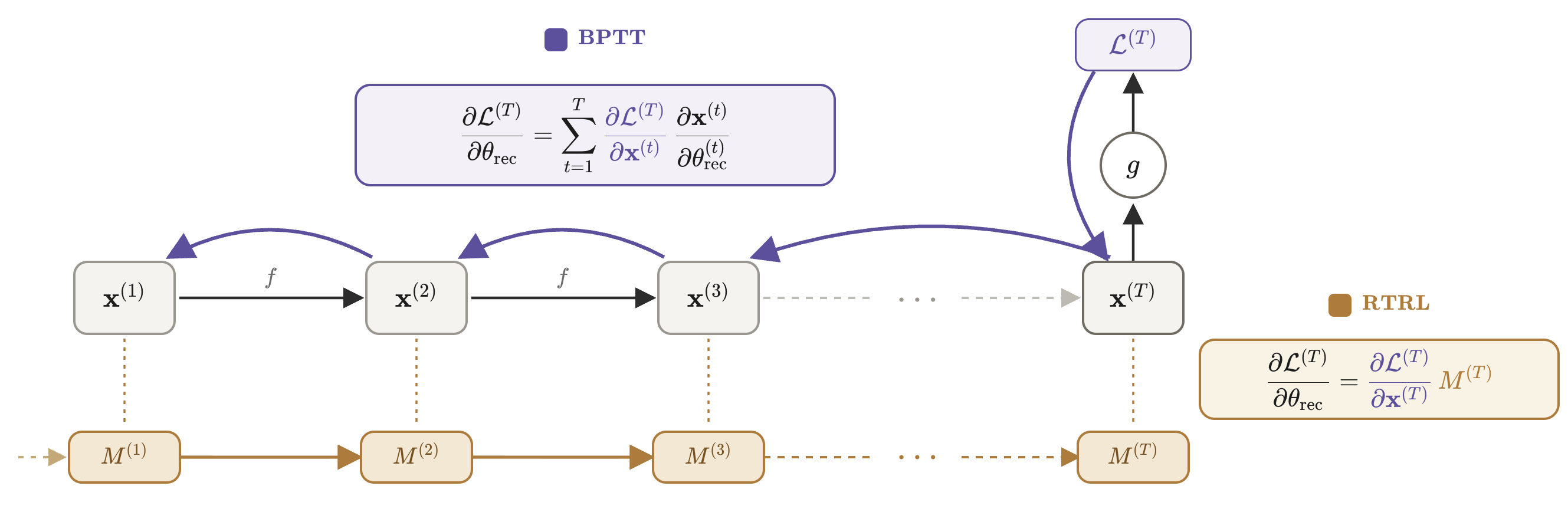}
    \caption{
    Comparison of temporal credit assignment in backpropagation-through-time
    (BPTT) and Real-Time Recurrent Learning (RTRL). BPTT computes recurrent
    parameter gradients by propagating loss derivatives backward through an
    unrolled state trajectory. RTRL instead maintains an online influence
    matrix during the forward execution of the network, allowing the
    current loss gradient to incorporate the historic influence of recurrent
    parameters without a reverse traversal of the full sequence. RTRL does not change how spatial credit assignment is performed.
    }
    \label{fig:bptt_rtrl_comparison}
\end{figure*}

\subsection{Real-Time Recurrent Learning}
\label{sec:rtrl}

\citet{williams1989learning} introduced Real-Time Recurrent Learning (RTRL) as an alternative to backpropagation-through-time (BPTT) for recurrent neural networks. Unlike BPTT, RTRL does not require the network to be unrolled through the full input sequence, and its memory cost is therefore independent of sequence length. Instead, temporal credit assignment is performed online by maintaining an influence matrix that tracks how the current hidden state depends on the recurrent parameters.

Figure~\ref{fig:bptt_rtrl_comparison} highlights the key differences between BPTT and RTRL. BPTT must store the whole latent-state trajectory, and computes parameter updates by stepping backwards through the unrolled computation graph. RTRL instead maintains an online influence matrix that captures the effect successive applications of the recurrent parameters have had on the current hidden state. This allows RTRL to compute exact recurrent parameter gradients without needing to store or traverse the full latent-state trajectory. RTRL trades the cost of maintaining this influence matrix for the cost of storing the full latent-state trajectory. 

Formally, this quantity is the sensitivity of the current hidden state to the recurrent parameters. For a single recurrent hidden state $\xx^{(t)}$ with recurrent parameters $\theta_{\mathrm{rec}}$, we define the influence matrix $\mathbf{M}^{(t)}$ as
\begin{equation}
\mathbf{M}^{(t)}
=
\frac{\partial \xx^{(t)}}{\partial \theta_{\mathrm{rec}}}.
\end{equation}
Applying the chain rule yields the standard RTRL recursion
\begin{align}
\mathbf{M}^{(t)}
&=
\frac{\partial \xx^{(t)}}{\partial \theta_{\mathrm{rec}}^{(t)}}
+
\frac{\partial \xx^{(t)}}{\partial \xx^{(t-1)}}
\frac{\partial \xx^{(t-1)}}{\partial \theta_{\mathrm{rec}}}
\notag\\
&=
\underbrace{\bar{\mathbf{M}}^{(t)}}_{\text{immediate influence}}
+
\underbrace{\mathbf{J}^{(t)}\mathbf{M}^{(t-1)}}_{\text{historic influence}},
\label{eq:rtrl_update_rule}
\end{align}
where $\bar{\mathbf{M}}^{(t)}$ is the immediate influence of $\theta_{\mathrm{rec}}$ on the current hidden state and $\mathbf{J}^{(t)} = \partial \xx^{(t)} / \partial \xx^{(t-1)}$ is the Jacobian of the hidden-state transition. Once $\mathbf{M}^{(t)}$ is available, the exact gradient of any instantaneous loss $\mathcal{L}^{(t)}$ with respect to the recurrent parameters can be written as
\begin{equation}
\label{eq:rtrl_bp}
\frac{\partial \mathcal{L}^{(t)}}{\partial \theta_{\mathrm{rec}}}
=
\frac{\partial \mathcal{L}^{(t)}}{\partial \xx^{(t)}}\mathbf{M}^{(t)}.
\end{equation}
Thus, unlike one-step updates that only account for the current application of the recurrent parameters, RTRL accumulates their full recursive influence through the hidden-state trajectory. Note that Eq.~\eqref{eq:rtrl_bp} yields the same gradient as BPTT, just computed differently.

The main drawback of RTRL relates to computational cost. For a dense recurrent layer with $n$ hidden units and $P = O(n^2)$ recurrent parameters, the influence matrix has size $n \times P = O(n^3)$ per layer. In multilayer recurrent networks the situation is more severe as each hidden layer depends on the parameters of all recurrent layers below it, and a naive $L$-layer construction requires $L(L+1)/2$ influence matrices \cite{irie2024exploringpromiselimitsrealtime}. Exact RTRL is therefore rarely used in large dense architectures. Several works make RTRL tractable through low-rank, factorised or sparse approximations to the influence matrix, trading exactness of $M^{(t)}$ for reduced computational cost and smaller memory requirements \cite{mujika2018approximatingrealtimerecurrentlearning,tallec2017unbiasedonlinerecurrentoptimization,menick2020practicalsparseapproximationreal}. Another approach is to use element-wise recurrent models, where each hidden unit depends only on its own recurrent parameters \cite{irie2024exploringpromiselimitsrealtime,zucchet2023onlinelearninglongrangedependencies}. In this setting, the influence matrix becomes block-sparse and $M^{(t)}$ can be maintained without needing any approximation, whilst model performance often remains competitive with dense counterparts.

\section{Temporal Predictive Coding with Real-Time Recurrent Learning}
\begin{figure*}[t]
\centering
\includegraphics[width=\textwidth]{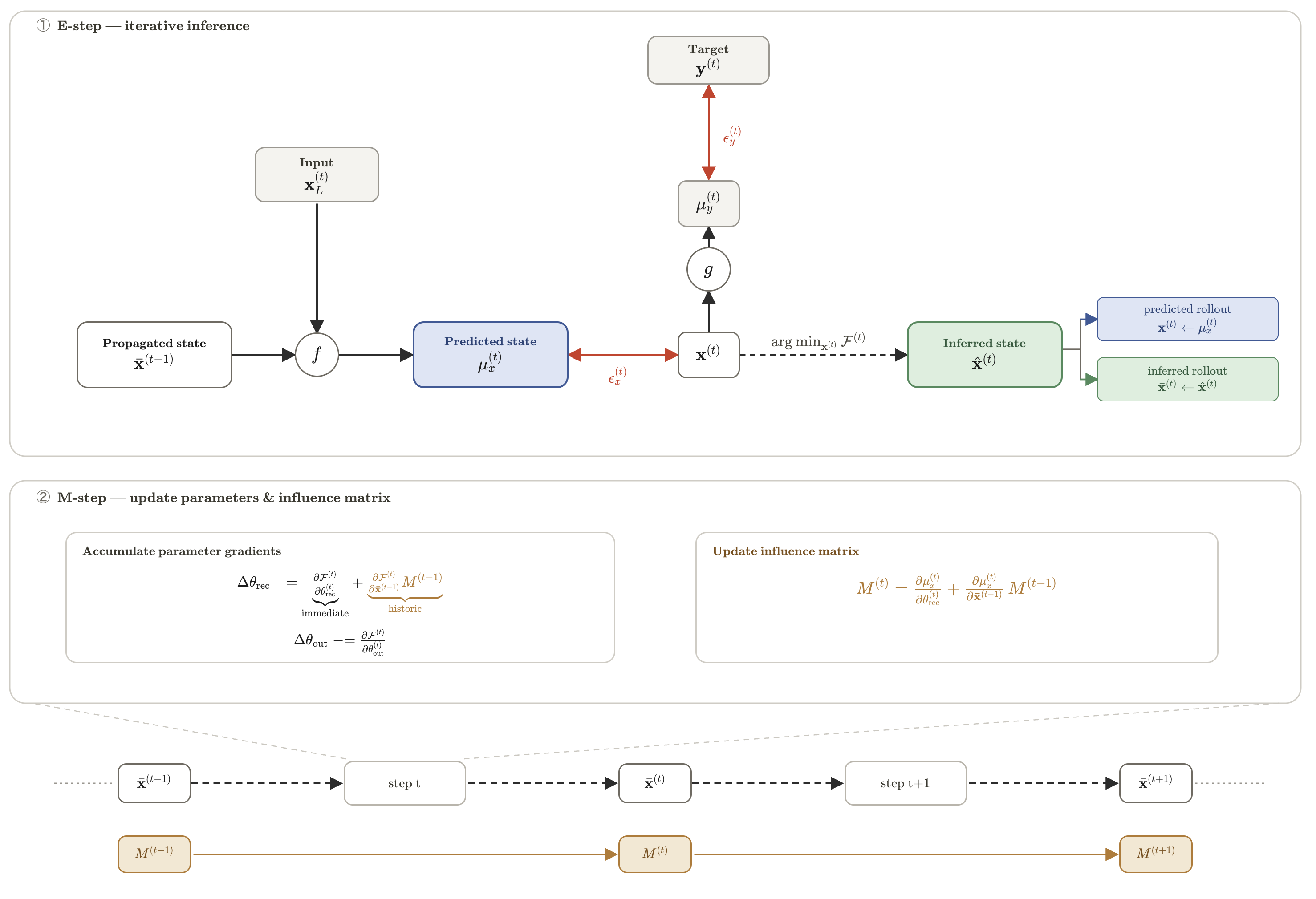}
\caption{
Overview of standard tPC and tPC-RTRL at a single timestep.
The upper panel shows the shared tPC E-step. Here, the recurrent prior
$f$ produces a \textcolor{myblue}{predicted} latent state (\textcolor{myblue}{blue}) for the current timestep; iterative inference then produces the \textcolor{mygreen}{inferred} state (\textcolor{mygreen}{green}) by minimising local \textcolor{red}{prediction errors} (\textcolor{myred}{red}). The middle panel shows the corresponding M-step. Standard tPC uses only the immediate component of the update, whereas tPC-RTRL additionally maintains an online \textcolor{mygold}{influence matrix} that includes its historic contribution to the current hidden state. \textcolor{mygold}{Gold} elements denote components introduced by tPC-RTRL. The lower panel illustrates the forward
propagation of the latent state and influence matrix across timesteps.
}
\label{fig:tpcrtrl_diagram}
\end{figure*}
\label{sec:method}
The previous section highlighted a structural limitation of tPC where the recurrent parameter update captures only the immediate application of the recurrent dynamics. In this section, we introduce Temporal Predictive Coding with Real-Time Recurrent Learning (tPC-RTRL), which augments tPC with an online influence matrix that aims to restore long-range temporal credit assignment whilst preserving the local structure of PC. 

Figure~\ref{fig:tpcrtrl_diagram} provides an overview of the method, contrasting it with tPC. The upper panel shows the standard
tPC inference computation shared by tPC and tPC-RTRL. The recurrent
prior predicts the current latent state, and iterative inference
adjusts that state to reduce local prediction errors. The middle panel
shows the difference in the recurrent M-step. Standard tPC uses only
the immediate parameter contribution, whereas tPC-RTRL introduces an influence matrix to carry historic parameter effects forward
through time. The lower panel shows the two quantities propagated
between successive timesteps.

\subsection{Predicted, inferred, and propagated latent states}
\label{sec:three_states}

To read Figure~\ref{fig:tpcrtrl_diagram}, it is useful to distinguish
three latent-state quantities with different computational roles. The
state $\bar{\xx}^{(t-1)}$ is the propagated latent state received
from the previous timestep. Given the current input and this propagated
state, the recurrent prior $f$ produces the predicted
latent state
\begin{equation}
\label{eq:predicted_state}
\mu_x^{(t)}
=
f(\xx_L^{(t)}, \bar{\xx}^{(t-1)}; \theta_{\mathrm{rec}}).
\end{equation}
The E-step then uses iterative inference to produce the
inferred latent state $\hat{\xx}^{(t)}$, which
reduces the instantaneous free energy by balancing the recurrent
prior prediction with the current output observation. The associated
prediction errors provide the local signals used for
parameter updates.

The distinction between $\mu_x^{(t)}$, $\hat{\xx}^{(t)}$, and
$\bar{\xx}^{(t)}$ is central to the method. The predicted state is the
feedforward recurrent transition, the inferred state is its
observation-corrected counterpart, and the propagated state determines
which of these quantities becomes the input to the next recurrent
transition. In Figure~\ref{fig:tpcrtrl_diagram}, blue denotes the
predicted state, green denotes the inferred state and its optional
propagation path, red denotes local prediction errors, and gold denotes
the additional temporal-influence quantities introduced by tPC-RTRL.

A practically important design choice is therefore whether the
propagated state is the inferred state,
$\bar{\xx}^{(t)} = \hat{\xx}^{(t)}$, or the recurrent prediction,
$\bar{\xx}^{(t)} = \mu_x^{(t)}$. The two alternatives are shown at the
right of the upper panel in Figure~\ref{fig:tpcrtrl_diagram}. The first option, choosing the inferred state and setting
$\bar{\xx}^{(t)} = \hat{\xx}^{(t)}$, ensures that each recurrent transition is conditioned on the latest latent correction. This may be natural in filtering-style settings, where new observations are available online at test-time and could be used to refine the internal state. However, when training autoregressive sequence models, it may not make sense to propagate $\hat{\xx}^{(t)}$. By propagating this corrected latent state when training, we are conditioning future dynamics on corrected states rather than the model's own autonomous predictions. In this regime, inference can continually repair the latent trajectory, reducing pressure on the recurrent dynamics to learn good open-loop rollouts, leading to poor performance when corrections are not available at test-time.

In the settings considered in the main text, predictive performance is
the primary objective. We therefore use the predictive-rollout
choice
\begin{equation}
\bar{\xx}^{(t)} = \mu_x^{(t)},
\end{equation}
and use the inferred state only to generate local prediction errors and
parameter updates. This is the blue rollout branch in
Figure~\ref{fig:tpcrtrl_diagram}, and it is the regime implemented by
Algorithm~\ref{alg:tpcrtrl}. The inferred-propagation alternative is discussed in \ref{app:inferred-propagation}.

\subsection{tPC-RTRL}
\label{sec:tpcrtrl_derivation}
The middle and lower panels of Figure~\ref{fig:tpcrtrl_diagram} show
the additional computation introduced by tPC-RTRL. Standard tPC already
provides the local inference dynamics in the upper panel, but its
recurrent update accounts only for the immediate effect of
$\theta_{\mathrm{rec}}$ on the current prediction. tPC-RTRL augments
this update with an online influence matrix that
stores how recurrent parameters have affected the latent state
propagated through earlier timesteps.

To account for the missing temporal dependency in Eq.~\eqref{eq:tpc_immediate_update}, we introduce an influence matrix
\begin{equation}
\mathbf{M}^{(t)}
=
\frac{\partial \bar{\xx}^{(t)}}{\partial \theta_{\mathrm{rec}}},
\end{equation}
which tracks how the recurrent parameters influence the propagated latent state. The recurrent parameter update can then be augmented to include both immediate and historic contributions:
\begin{align}
\Delta \theta_{\mathrm{rec}}
=
-
\left(
\underbrace{
\frac{\partial \F^{(t)}}{\partial \theta_{\mathrm{rec}}^{(t)}}
}_{\text{immediate}}
+
\underbrace{
\frac{\partial \F^{(t)}}{\partial \bar{\xx}^{(t-1)}}
\mathbf{M}^{(t-1)}
}_{\text{historic}}
\right). 
\label{eq:tpc_rtrl_update}
\end{align}
This is the tPC analogue of the RTRL update rule. The immediate term captures the current use of the recurrent parameters, whilst the historic term propagates their effect forward through the latent-state trajectory. The propagation choice of Section~\ref{sec:three_states} determines whether the recursion for $\mathbf{M}^{(t)}$ is exact or an approximation to the true influence of $\theta_{\mathrm{rec}}$ on the current latent state. We discuss both regimes below.

\paragraph{Predictive-rollout regime (exact recursion)} When $\bar{\xx}^{(t)} = \mu_x^{(t)}$, the influence update is exact:
\begin{equation}
\mathbf{M}^{(t)}
=
\frac{\partial \mu_x^{(t)}}{\partial \theta_{\mathrm{rec}}^{(t)}}
+
\frac{\partial \mu_x^{(t)}}{\partial \bar{\xx}^{(t-1)}}\mathbf{M}^{(t-1)}.
\label{eq:tpcrtrl_exact}
\end{equation}
\label{sec:predictive_rollout}
In this regime, tPC-RTRL coincides with exact RTRL on the recurrent latent trajectory executed by the feedforward components of the model. All experiments in the main text of this paper use this regime.

\paragraph{Inferred-propagation regime (approximate recursion)} When $\bar{\xx}^{(t)} = \hat{\xx}^{(t)}$, the transition from $\bar{\xx}^{(t-1)}$ to $\bar{\xx}^{(t)}$ is defined implicitly by iterative inference rather than a closed-form recurrence, and the exact influence recursion is no longer available. In this case there are two natural approximations. One can continue to update the influence matrix using Eq.~\eqref{eq:tpcrtrl_exact}, or alternatively evaluate the recursion using derivatives of the recurrent prediction at the converged inferred state:
\begin{equation}
\mathbf{M}^{(t)}
\approx
\left[
\frac{\partial \mu_x^{(t)}}{\partial \theta_{\mathrm{rec}}^{(t)}}
+
\frac{\partial \mu_x^{(t)}}{\partial \bar{\xx}^{(t-1)}}
\mathbf{M}^{(t-1)}
\right]_{\mu_x^{(t)} = \hat{\xx}^{(t)}}.
\label{eq:tpcrtrl_influence}
\end{equation}
For some models, Eqs.~\eqref{eq:tpcrtrl_exact} and \eqref{eq:tpcrtrl_influence} may be exactly the same. For example, if the model uses a linear projection as its transition function, these derivatives do not depend on the value of the current hidden state. A discussion of tPC-RTRL in this inferred-propagation regime is given in \ref{app:inferred-propagation} and Section~\ref{sec:beyond_predictive}.

\begin{algorithm*}[t]
\caption{One training step of tPC-RTRL in the predictive-rollout regime
(the blue rollout branch in Figure~\ref{fig:tpcrtrl_diagram}).}
\label{alg:tpcrtrl}
\begin{algorithmic}[1]
\Require Sequence $(\xx_L^{(1)}, \dots, \xx_L^{(T)})$ with targets $(\yy^{(1)}, \dots, \yy^{(T)})$; parameters $\theta_{\mathrm{rec}}, \theta_{\mathrm{out}}$; learning rate $\eta$.
\State Initialise hidden state $\bar{\xx}^{(0)}$; set influence matrix $\mathbf{M}^{(0)} \gets \partial \bar{\xx}^{(0)} / \partial \theta_{\mathrm{rec}}$.
\State Initialise gradient accumulators $\Delta \theta_{\mathrm{rec}} \gets 0,\; \Delta \theta_{\mathrm{out}} \gets 0$.
\For{$t = 1, \dots, T$}
    \State $\mu_x^{(t)} \gets f(\xx_L^{(t)}, \bar{\xx}^{(t-1)}; \theta_{\mathrm{rec}})$ \Comment{predict}
    \State Run iterative inference to obtain $\hat{\xx}^{(t)}$ minimising $\F^{(t)}$ \Comment{E-step}
    \State $\mathbf{M}^{(t)} \gets \partial \mu_x^{(t)} / \partial \theta_{\mathrm{rec}}^{(t)} + (\partial \mu_x^{(t)} / \partial \bar{\xx}^{(t-1)})\, \mathbf{M}^{(t-1)}$ \Comment{exact recursion}
    \State Accumulate recurrent grad: $\Delta \theta_{\mathrm{rec}} \mathrel{-}= \partial \F^{(t)}/\partial \theta_{\mathrm{rec}}^{(t)} + (\partial \F^{(t)}/\partial \bar{\xx}^{(t-1)})\, \mathbf{M}^{(t-1)}$ \Comment{M-step (recurrent)}
    \State Accumulate readout grad: $\Delta \theta_{\mathrm{out}} \mathrel{-}= \partial \F^{(t)} / \partial \theta_{\mathrm{out}}$ \Comment{M-step (readout)}
    \State $\bar{\xx}^{(t)} \gets \mu_x^{(t)}$ \Comment{predictive-rollout}
\EndFor
\State $\theta_{\mathrm{rec}} \gets \theta_{\mathrm{rec}} + \eta\, \Delta \theta_{\mathrm{rec}}$;\quad $\theta_{\mathrm{out}} \gets \theta_{\mathrm{out}} + \eta\, \Delta \theta_{\mathrm{out}}$
\end{algorithmic}
\end{algorithm*}

Importantly, the tPC-RTRL modifications affect only the recurrent learning rule. The inference dynamics remain exactly those of tPC, and non-recurrent parameters continue to use the standard tPC update, descending the gradient of the instantaneous free energy with respect to their current application, as this captures their full contribution:
\begin{equation}
\Delta \theta_{\mathrm{out}}
=
-\frac{\partial \F^{(t)}}{\partial \theta_{\mathrm{out}}^{(t)}}.
\label{eq:tpc_rtrl_readout_update}
\end{equation}
tPC-RTRL therefore preserves the local nature of PC, whilst augmenting the recurrent weight update with an online estimate of long-range temporal influence.

\subsection{Algorithmic summary}
\label{sec:algorithm_summary}
The recurrent parameter update of Eq.~\eqref{eq:tpc_rtrl_update} augments the standard tPC rule with an online estimate of long-range temporal influence, whilst leaving the inference dynamics and non-recurrent parameter updates unchanged. Throughout this work, we adopt the forward-update convention of \citet{qi2025training} where parameter updates move the feedforward predictions towards the converged values. This choice keeps parameter updates anchored to the feedforward trajectory the model follows at test-time and has been shown to improve performance. Realising the forward-update convention requires a small amount of additional bookkeeping to implement, which we detail in \ref{app:tpcrtrl_algorithm}. We leave the notation in the main text unchanged to keep the presentation of the algorithm as clear as possible and in keeping with standard PC and tPC definitions.

In practice, the recurrent gradients in Eq.~\eqref{eq:tpc_rtrl_update} may be applied online at each timestep, or accumulated across a sequence and applied in a time-batched manner. We adopt the latter, which avoids issues with stale gradients and is more convenient in minibatch training.

Algorithm~\ref{alg:tpcrtrl} is the pseudocode counterpart of the
predictive-rollout branch in
Figure~\ref{fig:tpcrtrl_diagram}. At each timestep, the recurrent prior
first computes the predicted state $\mu_x^{(t)}$. Iterative inference
then produces $\hat{\xx}^{(t)}$ and the local prediction errors required
for learning, without replacing the state propagated by the recurrent
dynamics. In parallel, the influence matrix $\mathbf{M}^{(t)}$ is
updated forward in time. The recurrent gradient accumulator receives
both the immediate contribution from the current transition and the
historic contribution carried by $\mathbf{M}^{(t-1)}$, while the
predicted state $\mu_x^{(t)}$ is propagated to the next timestep.

\subsection{Relation to Backpropagation-Through-Time}
\label{sec:bptt_relation}
PC's relationship to BP has been a central thread in the recent PC literature. A series of results have shown that PC can approximate BP gradients in static feedforward networks \cite{Whittington2017AnAO,DBLP:journals/corr/abs-2006-04182} and, under explicit assumptions, recover them exactly across general computation graphs \cite{NEURIPS2020_fec87a37,salvatori2023reversedifferentiationpredictivecoding}. Beyond helping to clarify what PC may be doing as an optimisation algorithm, these equivalences demonstrate that the spatial locality of PC does not come at the cost of the gradient quality that has made BP so successful.

Here we extend these proofs to tPC-RTRL. In the predictive-rollout regime introduced in Section~\ref{sec:predictive_rollout}, the RTRL recursion is exact and the tPC-RTRL recurrent update recovers BPTT gradients under explicit assumptions on inference and the readout pathway. This shows that the locality properties demanded by neuromorphic hardware (Section~\ref{sec:neuromorphic_motivation}) can be satisfied in both space and time without sacrificing the optimisation quality of BPTT.

\begin{proposition}[tPC-RTRL recovers BPTT in the predictive-rollout regime]
\label{prop:tpcrtrl_bptt}
Assume (i) inference converges at each timestep, (ii) the
fixed-prediction assumption \cite{DBLP:journals/corr/abs-2006-04182}
holds for the readout network, and (iii) the propagated state
follows the predictive-rollout $\bar{\xx}^{(t)} = \mu_x^{(t)}$. Then
the tPC-RTRL recurrent updates, summed over a sequence of length
$T$, recover the exact BPTT gradient with respect to the recurrent
parameters $\theta_{\mathrm{rec}}$:
\begin{equation}
\sum_{t=1}^{T}
\left[
\frac{\partial \F^{(t)}}{\partial \theta_{\mathrm{rec}}^{(t)}}
+
\frac{\partial \F^{(t)}}{\partial \bar{\xx}^{(t-1)}}
\mathbf{M}^{(t-1)}
\right]
=
\frac{\partial}{\partial \theta_{\mathrm{rec}}}
\sum_{t=1}^{T}\ell_t,
\label{eq:prop_bptt_match}
\end{equation}
where $\mathbf{M}^{(t)} = \partial \bar{\xx}^{(t)} / \partial \theta_{\mathrm{rec}}$
is the RTRL influence matrix.
\end{proposition}

The argument combines two points. A spatial PC
equivalence at the recurrent state gives $\epsilon_x^{(t)} =
-\partial \ell_t / \partial \bar{\xx}^{(t)}$ at inference
equilibrium, so the PC error acts as the backpropagated signal
through the non-recurrent layers. The exact RTRL recursion on the propagated
trajectory (Eq. \ref{eq:tpcrtrl_exact}) then provides the historic terms of the
tPC-RTRL update. Summing over
time gives the BPTT gradient. The full derivation is in
\ref{app:bptt_equivalence}.

\paragraph{Scope of the equivalence and our experimental regime} Proposition~\ref{prop:tpcrtrl_bptt} identifies an operating point at which tPC-RTRL coincides exactly with BPTT. Its role in this paper is to establish that the BPTT gradient lies within the family of updates tPC-RTRL can produce. 

However, our experiments deliberately do not enforce this regime. We do this for a few reasons. First, guaranteeing convergence at each timestep, at least to the precision required for the theoretical BPTT equivalence, is likely to be difficult in practice, and would incur a computational cost that is ideally avoided. Furthermore, the fixed-prediction assumption \cite{DBLP:journals/corr/abs-2006-04182} requires us to alter the inference dynamics, evaluating all derivatives during inference at their feedforward values. This would mean that inference is no longer performing gradient descent on the defined free energy, but instead on a slightly modified objective, and would, more fundamentally, mean that the inferred latent state no longer corresponds to an approximation of the model's posterior over the latent. Under the variational interpretation of PC, the E-step performs approximate Bayesian inference of the latent given the current observation and previous inferred state \cite{Friston2003,Millidge2023.05.15.540906}. The filtering procedure of Section~\ref{sec:filtering} exploits this property directly, clamping true-state observations at deployment time and allowing iterative inference to revise the latent trajectory in a way that is consistent with the generative model. Inference dynamics tuned to recover BPTT gradients would no longer necessarily admit this interpretation, and might not allow for online state correction at test-time.

Furthermore, a growing body of work suggests that PC's behaviour outside the equivalence regime is itself a source of its value, with reports of improved sample efficiency, robustness, and updates that capture aspects of higher-order optimisation \cite{song_inferring_2024,mali2024tightstabilityconvergencerobustness}. The empirical question we therefore ask in Section~\ref{sec:results} is not whether tPC-RTRL can exactly match BPTT under tightly controlled assumptions, but whether, when allowed to operate outside of this controlled regime, it produces recurrent updates of comparable optimisation quality. We return to this question in Section~\ref{sec:near_equivalence}.

\subsection{Instantiation on the RG-LRU}
\label{sec:rglru}

The experiments in Sections~\ref{sec:wikitext}--\ref{sec:nanodrone} apply tPC-RTRL to a Real-Gated Linear Recurrent Unit (RG-LRU) \cite{de2024griffinmixinggatedlinear}. We use this architecture because its recurrence is element-wise in the hidden dimension, which makes the exact RTRL recursion of Eq.~\eqref{eq:tpcrtrl_exact} tractable. Each hidden unit depends only on its own row of parameters, so no cross-unit influences need to be tracked. Here we provide the cell definition, the per-timestep free energy used during inference, the element-wise RTRL recursion, and the resulting recurrent gradient. Further information including the regression-output variant used on the nanodrone model, and the optional input-projection and learnt state-initialisation heads is deferred to \ref{app:rglru}.

For clarity, the tPC and tPC-RTRL formulations of Sections~\ref{sec:tpc_background}--\ref{sec:tpcrtrl_derivation} were stated with a single readout layer, so that the recurrent latent $\xx^{(t)}$ directly predicts the output by $g(\,\cdot\,;\theta_{\mathrm{out}})$. The RG-LRU models we use in the presented experiments have multiple readout layers. In this setting, our free energy formulation changes to include a latent state at each layer. For example, Eq.~\eqref{eq:rglru_free_energy} minimises the free energy jointly over the recurrent latent $\xx^{(t)}$ and an intermediary readout latent $\mathbf{o}^{(t)}$. This is a straightforward extension of the single-layer case, where we are effectively doing tPC over the recurrent latents, and PC over the readout latents as they lack the recurrent prior. This means that all layers, both recurrent and non-recurrent, are updated according to layer-local rules, regardless of the depth of the readout stack.

\paragraph{RG-LRU transition} Writing $\xx_L^{(t)} \in \mathbb{R}^{I}$ for the input to the cell and $\bar{\xx}^{(t-1)} \in \mathbb{R}^{H}$ for the propagated recurrent state from the previous timestep, the recurrent prediction $\mu_x^{(t)} \in \mathbb{R}^{H}$ is given by
\begin{align}
g_a^{(t)} &= \sigma(W_a \xx_L^{(t)} + b_a), \\
g_z^{(t)} &= \sigma(W_z \xx_L^{(t)} + b_z), \\
\log a^{(t)} &= c\, g_a^{(t)} \odot \log\sigma(\Lambda), \\
\gamma^{(t)} &= \sqrt{1 - \exp(2 \log a^{(t)})}, \\
\mu_x^{(t)} &= a^{(t)} \odot \bar{\xx}^{(t-1)} + \gamma^{(t)} \odot \bigl(g_z^{(t)} \odot \xx_L^{(t)}\bigr),
\label{eq:rglru_transition}
\end{align}
with recurrent parameters $\theta_{\mathrm{rec}} = \{\Lambda, W_a, b_a, W_z, b_z\}$, a fixed scalar $c$, and $\Lambda \in \mathbb{R}^{H}$. The two gates $g_a^{(t)}, g_z^{(t)} \in (0,1)^H$ modulate the decay $a^{(t)}$ and the input pathway respectively, and the recurrence is element-wise in the hidden dimension as each coordinate of $\mu_x^{(t)}$ depends only on the corresponding coordinate of $\bar{\xx}^{(t-1)}$. When the input dimensionality differs from the hidden dimensionality, an additional linear projection $W_{\mathrm{in}}, b_{\mathrm{in}}$ is inserted in front of $\xx_L^{(t)}$ in the $g_z^{(t)} \odot \xx_L^{(t)}$ product; we keep this implicit in the main text and defer the details to \ref{app:rglru}.

\paragraph{Free energy and inference} The RG-LRU is wrapped in a ReLU readout layer and a linear output head. During inference, let $\xx^{(t)}$ denote the inferred recurrent latent, $\mathbf{o}^{(t)}$ the inferred readout latent, $\yy^{(t)}$ the target, and $\phi$ the ReLU. With Gaussian conditionals at the recurrent and readout states and a softmax conditional at the output \cite{pinchetti2022predictivecodinggaussiandistributions}, the per-timestep free energy is
\begin{align}
\F^{(t)}
=&\;
\frac{1}{2 H}\,\|\xx^{(t)} - \mu_x^{(t)}\|_2^2 \notag\\
&+
\frac{1}{2 R}\,\|\mathbf{o}^{(t)} - \phi(W_r \xx^{(t)} + b_r)\|_2^2 \notag\\
&+
\frac{1}{C}\,\ell_{\mathrm{CE}}\!\bigl(W_l\,\mathbf{o}^{(t)} + b_l,\; \yy^{(t)}\bigr),
\label{eq:rglru_free_energy}
\end{align}
where $W_r, b_r$ are the readout weights, $W_l, b_l$ are the output-head weights, $R$ is the readout dimension, and $C$ is the output dimension. This gives readout parameters $\theta_{\mathrm{out}} = \{W_r, b_r, W_l, b_l\}$. The per-layer normalisations $1/H, 1/R, 1/C$ aim to keep the scale of each energy term independent of layer width. Inference minimises $\F^{(t)}$ jointly over $\{\xx^{(t)}, \mathbf{o}^{(t)}\}$ via a fixed number of steps of momentum-accelerated gradient descent initialised at the feedforward values $\xx^{(t)} \leftarrow \mu_x^{(t)}$, $\mathbf{o}^{(t)} \leftarrow \phi(W_r \mu_x^{(t)} + b_r)$. With the converged latent, $\hat{\xx}^{(t)}$, the associated recurrent prediction error is given by
\begin{equation}
\hat{\epsilon}_x^{(t)} = \hat{\xx}^{(t)} - \mu_x^{(t)}
\end{equation}
and supplies the local signal used by the recurrent update below. The explicit gradient forms for $\partial \F^{(t)} / \partial \xx^{(t)}$ and $\partial \F^{(t)} / \partial \mathbf{o}^{(t)}$, together with the inference step size and number of iterations used in each experiment, are given in \ref{app:rglru_inference}.

\paragraph{Element-wise RTRL recursion} Because the recurrent prediction in Eq.~\eqref{eq:rglru_transition} is element-wise in the hidden dimension, the Jacobian of $\mu_x^{(t)}$ with respect to the propagated state collapses to $\partial \mu_x^{(t)} / \partial \bar{\xx}^{(t-1)} = \mathrm{diag}(a^{(t)})$. For any recurrent parameter $\theta \in \theta_{\mathrm{rec}}$, let $\mathbf{M}_{\theta}^{(t)} = \partial \bar{\xx}^{(t)} / \partial \theta$, the exact recursion of Eq.~\eqref{eq:tpcrtrl_exact} then becomes
\begin{equation}
\mathbf{M}_{\theta}^{(t)}
=
a^{(t)} \odot \mathbf{M}_{\theta}^{(t-1)}
+
\frac{\partial \mu_x^{(t)}}{\partial \theta^{(t)}},
\label{eq:rglru_rtrl}
\end{equation}
where the chain term has reduced to element-wise multiplication by $a^{(t)}$ along the hidden axis. Since each hidden unit depends only on its own row of $\theta_{\mathrm{rec}}$, the influence tensors carry only a single hidden dimension, rather than a full $H \times H$ Jacobian, and the total per-example recurrent storage is $O(H + HI)$, as summarised in Section~\ref{sec:memory}.

\paragraph{Recurrent parameter update} In the predictive-rollout regime, the immediate and historic terms of Eq.~\eqref{eq:tpc_rtrl_update} combine through the influence tensor of Eq.~\eqref{eq:rglru_rtrl}, and the time-batched gradient accumulator $\Delta \theta_{\mathrm{rec}}$ of Algorithm~\ref{alg:tpcrtrl} takes the simple form
\begin{equation}
\Delta \theta
=
\frac{1}{H}\sum_{t=1}^{T} \hat{\epsilon}_x^{(t)} \cdot \mathbf{M}_{\theta}^{(t)}
\qquad \text{for each } \theta \in \theta_{\mathrm{rec}},
\label{eq:rglru_gradient}
\end{equation}
applied at the end of the sequence as $\theta \gets \theta + \eta\,\Delta \theta$. The inference dynamics, the RTRL recursion, and the readout-parameter updates are all applied layer-locally to the cell, with no global BPTT-style reverse traversal through time. This is the form of the tPC-RTRL update used in all RG-LRU experiments reported in Section~\ref{sec:results}.

\section{Results}
\label{sec:results}
Section~\ref{sec:tpc_background} identified a structural limitation of standard tPC, showing that its recurrent parameter update neglects the historic influence of the recurrent parameters along the latent-state trajectory, restricting the model's ability to assign credit across long temporal horizons. The experiments in this section are organised around this limitation. We begin with the copy task (Section~\ref{sec:copy}), which serves as a controlled diagnostic in which the failure of standard tPC, and the recovery offered by tPC-RTRL, can be cleanly isolated. We then ask whether this advantage persists in more realistic sequence-modelling problems and larger recurrent models, evaluating tPC-RTRL on byte-level language modelling (Section~\ref{sec:wikitext}), machine translation (Section~\ref{sec:translation}), and nanodrone system identification (Section~\ref{sec:nanodrone}). Finally, we present evidence that the iterative inference dynamics of tPC-RTRL admit a natural filtering interpretation at test-time (Section~\ref{sec:filtering}).

\subsection{Copy Task}
\label{sec:copy}
The copy task provides a controlled diagnostic of long-range temporal credit assignment, and cleanly separates methods that can effectively propagate gradient information across a temporal gap from those that cannot. We therefore use it to isolate the limitations of standard tPC identified in Section~\ref{sec:tpc_background}, and to show that tPC-RTRL's augmentation of the recurrent update with an influence matrix is sufficient to restore long-range temporal credit assignment and solve the task.

\begin{figure}[t]
\centering
\includegraphics[width=0.35\textwidth]{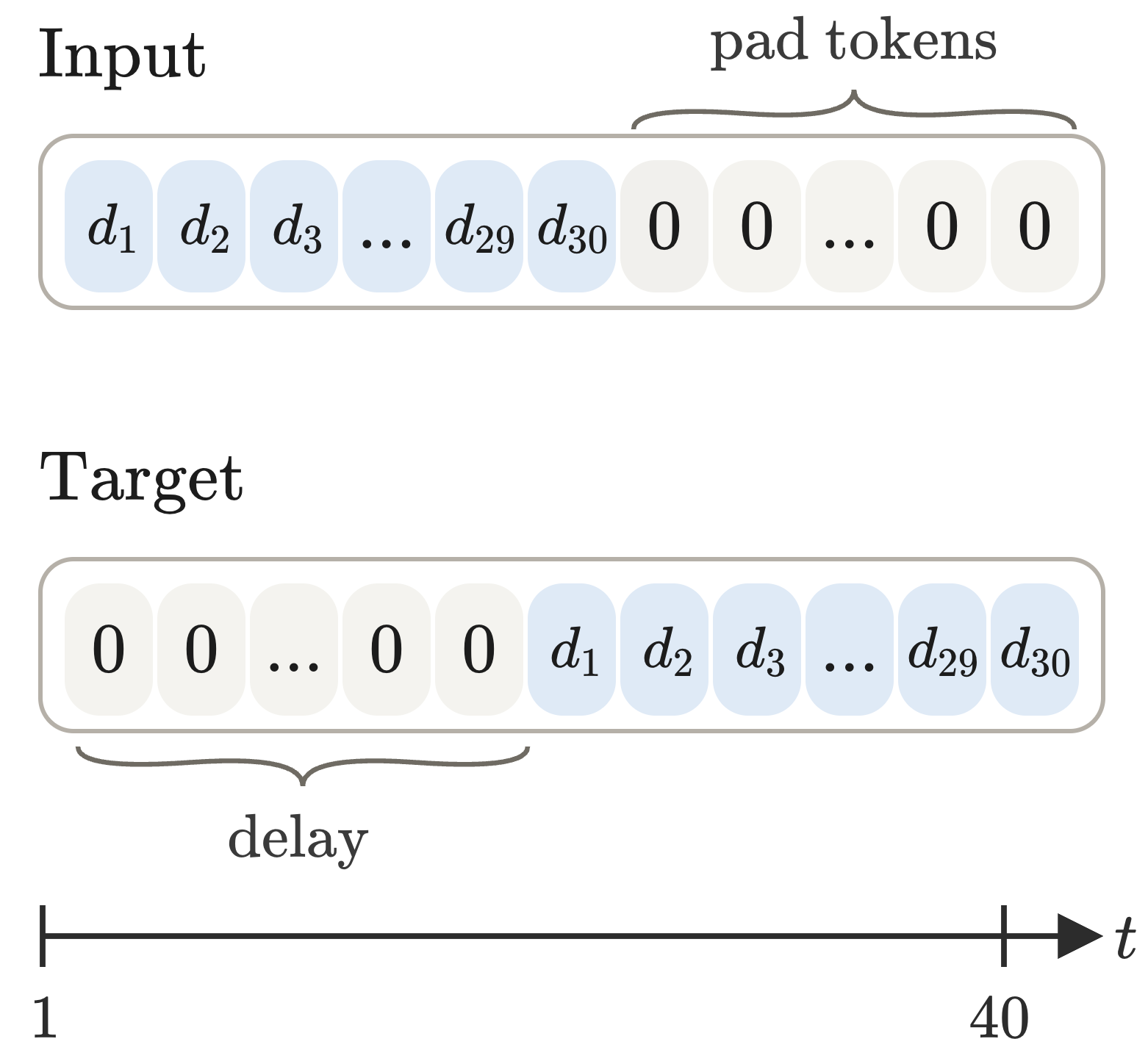}
\caption{
Copy task construction. A length-30 sequence of digits ($d_i$) is followed by padding in the input, while the target is the same sequence shifted into the future. $d_i$ are sampled uniformly from the integers 1--9 and one-hot encoded.
}
\label{fig:copy_task_schematic}
\end{figure}

\paragraph{Task definition}
Figure \ref{fig:copy_task_schematic} illustrates the copy task construction. The model is presented with a sequence of digits and must output the same sequence after a fixed delay. For this task, the information required to predict each output is separated from the learning signal by several timesteps, making it difficult for models to learn without a mechanism for propagating gradient information across this gap. The input is a sequence of 30 digits sampled uniformly from the integers 1--9, with 0 reserved as a padding token. The target is the same sequence translated 10 timesteps into the future, with both sequences padded to a total of 40 timesteps.

\paragraph{Experimental setup}
All models use a single-layer recurrent sequence model with 128 $\tanh$ hidden units and a per-timestep output head. Four models are trained on this task, one with each learning algorithm discussed in this work: tPC-RTRL, tPC, BPTT and spatial BP. Spatial BP represents a form of truncated BPTT where only local temporal information is used to calculate recurrent parameter gradients, detaching the hidden state at each timestep. All models are trained with a batch size of 16 and are evaluated on 200 randomly generated test samples every 16 batches. Reported values are means and standard deviations over 5 seeds, with the same seeds used for all methods. For full experimental details, see \ref{app:copy_task_setup}.

\paragraph{Results}
Table~\ref{tab:copy_task_results} shows a clear separation between methods that propagate recurrent credit through time and those that do not. BPTT and tPC-RTRL solve the task near-perfectly (accuracies of $99.99 \pm 0.01$ and $99.99 \pm 0.02$ respectively), achieving mean train and validation metrics matching to three significant figures. In contrast, spatial BP and standard tPC converge to a substantially worse solution, with validation accuracy of around 40\%. Overall, standard tPC closely matches spatial BP, whilst tPC-RTRL closely matches BPTT.

In this small and controlled setting, we are able to tune the PC inference dynamics so that tPC-RTRL reproduces BPTT almost exactly, and standard tPC reproduces spatial BP. These results support the claim that the failure mode of standard tPC on long-horizon sequence tasks is specifically a failure of recurrent credit assignment. On the copy task the relevant supervision arrives only after a long delay. A one-step recurrent update therefore appears insufficient to bridge the gap between presentation of the relevant information and the supervision signal, and the model fails to reliably learn. By contrast, tPC-RTRL overcomes this limitation by explicitly propagating recurrent influence through time, allowing tPC to recover the same optimisation behaviour as BPTT on this task.

\begin{table*}[t]
    \centering
    \small
    \begin{tabular}{lcccccc}
        \toprule
        Method & Train loss ($\downarrow$) & Val loss ($\downarrow$) & Val acc.\ (\%) ($\uparrow$) \\
        \midrule
        Spatial BP & $1.658 \pm 0.044$ & $1.652 \pm 0.047$ & $39.76 \pm 1.79$ \\
        tPC       & $1.657 \pm 0.044$ & $1.652 \pm 0.049$ & $39.79 \pm 1.86$ \\
        BPTT      & $0.058 \pm 0.007$ & $0.055 \pm 0.007$ & $99.99 \pm 0.01$ \\
        tPC-RTRL  & $0.058 \pm 0.007$ & $0.055 \pm 0.008$ & $99.99 \pm 0.02$ \\
                \bottomrule
    \end{tabular}
    \caption{
    Copy task performance, reported as mean $\pm$ standard deviation over 5 seeds. tPC-RTRL matches BPTT, whilst standard tPC matches spatial BP. This shows that the decisive factor on this task is not PC itself, but whether recurrent credit is propagated across time.
    }
    \label{tab:copy_task_results}
\end{table*}

\subsection{Language Modelling on WikiText-103}
\label{sec:wikitext}

The results of Section~\ref{sec:copy} establish that tPC-RTRL addresses the long-range credit-assignment limitation of tPC in a controlled setting. We now test whether this persists in more realistic sequence-modelling problems and larger recurrent models. Here, we study byte-level language modelling on WikiText-103 \cite{merity2016pointersentinelmixturemodels}, and in Section~\ref{sec:translation}, machine translation on a subset of CCMatrix \cite{schwenk2020ccmatrixminingbillionshighquality}. In both experiments we use the Real-Gated Linear Recurrent Unit (RG-LRU) \cite{de2024griffinmixinggatedlinear} introduced in Section~\ref{sec:rglru}. We choose this recurrent module because its element-wise recurrence makes exact RTRL-style influence tracking tractable at the widths used here, while still providing a richer gated recurrent architecture than the model used in the copy task.

\paragraph{Task definition}
The corpus is converted directly to UTF-8 bytes and modelled as a sequence over a vocabulary of size 256. Training examples are formed by sampling, with replacement, random contiguous windows of length 257 from the training split. For each window, the first 256 bytes are used as inputs and the final 256 bytes are used as next-byte prediction targets. Validation and test evaluation use sequential windows constructed in the same way, slicing the whole dataset into input--output pairs without resampling.

\paragraph{Experimental setup}
Four models are trained on this task, one with each learning algorithm. The language model consists of a frozen pretrained byte embedding layer and the RG-LRU model introduced in Section~\ref{sec:rglru}. Each model has 1{,}313{,}536 total trainable parameters. Of these, 525{,}824 are recurrent parameters that require a temporal credit-assignment mechanism to ensure effective learning. Each model is trained for 400{,}000 training steps and evaluated every 5{,}000 steps on the validation set. The best checkpoint per seed is selected using validation bits-per-character (BPC) and used to report final test metrics. Reported values are means and standard deviations over 5 seeds. For full experimental details, including model layer sizes, see \ref{app:exp_details_wikitext}.

\begin{figure}[t]
    \centering
    \includegraphics[width=\linewidth]{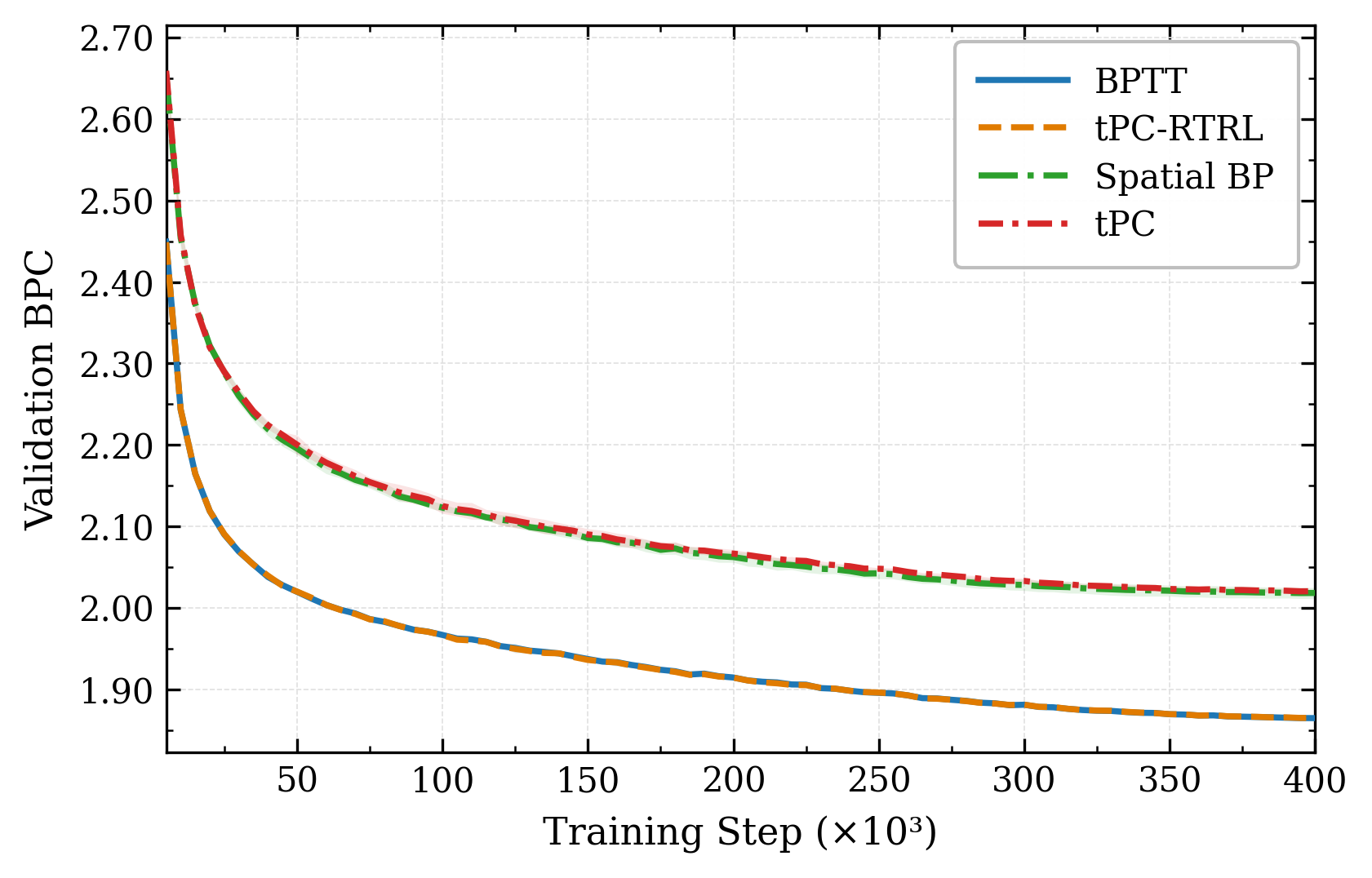}
    \caption{
    Validation bits-per-character on WikiText-103 as a function of training steps for BPTT, spatial BP, tPC, and tPC-RTRL. Lines show the mean over 5 seeds.
    }
    \label{fig:wikitext_bpc_curves}
\end{figure}

\paragraph{Results} Table~\ref{tab:wikitext_results} and Figure~\ref{fig:wikitext_bpc_curves} show that tPC-RTRL closes the gap to BPTT on byte-level language modelling, whilst standard tPC tracks the spatial BP baseline throughout training. BPTT reaches a mean validation BPC of 1.864, and tPC-RTRL reaches 1.865, a difference of 0.001 BPC that falls near seed variance. In contrast, tPC (2.020) and spatial BP (2.018) plateau above the BPTT/tPC-RTRL pair, and the two curves remain closely coupled across all 400{,}000 training steps (see Figure~\ref{fig:wikitext_bpc_curves}).

The correspondence observed on the copy task therefore persists at the scale of WikiText-103. The near-zero residual gap between tPC-RTRL and BPTT at this scale further suggests that, at least for this model size and dataset, any optimisation differences introduced by PC do not measurably degrade performance once temporal credit assignment is handled correctly.

\begin{table}[t]
    \centering
    \small
    \begin{tabular}{lccc}
        \toprule
        Method     & Best val.\ BPC ($\downarrow$) & Test BPC ($\downarrow$)\\
        \midrule
        Spatial BP & $2.018 \pm 0.006$ & $2.051 \pm 0.007$\\
        tPC        & $2.020 \pm 0.002$ & $2.054 \pm 0.004$\\
        BPTT       & $1.864 \pm 0.002$ & $1.895 \pm 0.001$\\
        tPC-RTRL   & $1.865 \pm 0.001$ & $1.895 \pm 0.002$\\
        \bottomrule
    \end{tabular}
    \caption{
        Byte-level language modelling results on WikiText-103.
        Values are reported as mean $\pm$ standard deviation over 5 random seeds.
        The best validation BPC per seed is used for model selection; test BPC is evaluated on the corresponding best checkpoint.
    }
    \label{tab:wikitext_results}
\end{table}

\subsection{Machine Translation on CCMatrix (English--French)}
\label{sec:translation}
 
Next, we look at machine translation, another sequence modelling problem which we consider a stronger test of temporal credit assignment. This is because the model must consume an entire source sentence before producing any target token, introducing a long-range dependency between the corresponding components of the source and target sentence, similar to that of the copy task in Figure \ref{fig:copy_task_schematic}. Algorithms that do not effectively propagate gradient information through time are therefore expected to struggle.

\paragraph{Task definition}
CCMatrix \cite{schwenk2020ccmatrixminingbillionshighquality} is a large-scale parallel corpus of 1.5 billion sentence pairs across 200 languages. As it has no canonical benchmark split and is far larger than our compute budget admits, we construct an English--French subset by sampling 600{,}000 sentence pairs, allocating 500{,}000 for training, 50{,}000 for validation, and 50{,}000 for testing. Sentences are filtered to a maximum combined (source + target) length of 128 tokens, lowercased, and tokenised with a Byte-Pair Encoding (BPE) tokeniser \cite{sennrich2016neural} trained jointly on source and target sentences from the training split, yielding a vocabulary of 10{,}000 tokens.

\paragraph{Experimental setup}
Four models are trained on this task, one with each learning algorithm. The model architecture mirrors the byte-level language model of Section~\ref{sec:wikitext}, now operating over BPE tokens. It consists of a frozen pretrained token embedding layer, and the RG-LRU model introduced in Section~\ref{sec:rglru}. All models contain 13{,}399{,}824 trainable parameters, of which 2{,}100{,}224 are recurrent parameters.

From epoch 5 onward we probabilistically drop target input tokens, increasing the drop probability throughout training. This aims to discourage teacher-forcing dependence and encourage the recurrent state to carry longer-range predictive information. Due to compute constraints we train one model per method and use the final checkpoint to report test metrics. Full details are presented in \ref{app:exp_details_translation}.

\begin{figure}[t]
    \centering
    \includegraphics[width=\linewidth]{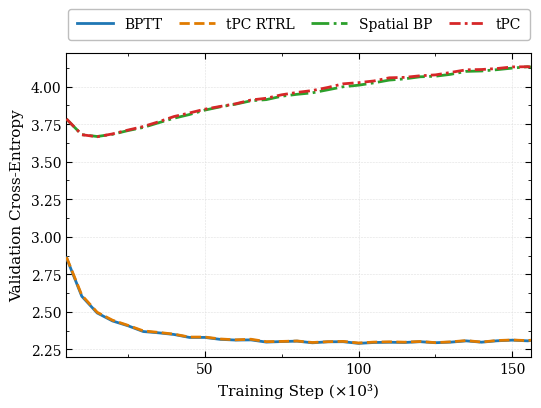}
    \caption{
     Validation cross-entropy loss (with label smoothing at 0.05) on the CCMatrix English--French translation task as a function of training steps for BPTT, spatial BP, tPC, and tPC-RTRL.
    }
    \label{fig:translation_curves}
\end{figure}

\paragraph{Results} Table~\ref{tab:translation_results} and Figure~\ref{fig:translation_curves} show that the observations seen on the copy task and WikiText-103 extend to machine translation. tPC-RTRL closely matches BPTT in both test perplexity and BLEU, whilst standard tPC tracks spatial BP throughout training. Note that the spatial BP and tPC curves show an increase in validation loss from epoch 5 onwards, suggesting the naive credit assignment mechanisms struggle to maintain performance under the input token dropout schedule. Translation appears to be a stronger test of long-range credit assignment than next-byte language modelling as the model must associate target tokens with source-side context separated by the full encoder-side rollout. The persistent gap between BPTT/tPC-RTRL and tPC/spatial-BP confirms that this dependency requires an effective temporal credit assignment mechanism, and that tPC-RTRL achieves this whilst relying only on local computations.

\begin{table}[t]
    \centering
    \small
    \begin{tabular}{lccc}
        \toprule
        Method     & Test PPL ($\downarrow$) & Test BLEU ($\uparrow$) \\
        \midrule
        Spatial BP & 24.56 & 4.58 \\
        tPC        & 24.53 & 4.75 \\
        BPTT       & 5.50 & 20.29 \\
        tPC-RTRL   & 5.52 & 20.23 \\
        \bottomrule
    \end{tabular}
    \caption{
        Perplexity (PPL) and BLEU score on the CCMatrix English--French subset. tPC-RTRL matches BPTT on both metrics, whilst standard tPC matches spatial BP. Metrics are a single run, with the final checkpoint used for evaluation.
    }
    \label{tab:translation_results}
\end{table}

\subsection{System Identification with tPC-RTRL}
\label{sec:nanodrone}
The experiments in Sections \ref{sec:copy}--\ref{sec:translation} have so far considered temporal credit assignment in discrete-token sequence tasks. To assess whether the BPTT/tPC-RTRL equivalence extends to continuous problems with nonlinear dynamics, we evaluate the four learning algorithms on a nanodrone system-identification benchmark \cite{Busetto_2026}. This section reports the nanodrone learning result; evidence for the filtering interpretation of tPC-RTRL, using the same dataset, is presented in Section~\ref{sec:filtering}.

\paragraph{Task definition}
\label{sec:nanodrone_setup}
We use the Nonlinear System Identification Nanodrone Benchmark \cite{Busetto_2026}, a real-flight dataset collected on a Crazyflie 2.1 nano-quadrotor (mass ${\approx}27$\,g, rotor span 92\,mm). The benchmark comprises aggressive flight trajectories recorded at 100\,Hz synchronising onboard motor commands with the drone's position, velocity, orientation, and angular velocity measured by an external motion-capture system, or onboard gyroscope.

The goal is to take a 4-dimensional sequence of propeller angular velocities $u_t \in \mathbb{R}^4$ as input and predict the 9-dimensional drone state
\begin{equation}
    s_t = \bigl[v^b_t,\; \phi_t,\; \omega_t\bigr] \in \mathbb{R}^9,
\end{equation}
where $v^b_t \in \mathbb{R}^3$ is the linear velocity expressed in the body frame, $\phi_t \in \mathbb{R}^3$ is the orientation encoded as a rotation vector, and $\omega_t \in \mathbb{R}^3$ is the angular velocity. Position is excluded from the direct prediction target and is recovered by integrating the predicted world-frame velocity, as is common in these types of task where position is typically unbounded.

\paragraph{Experimental setup}
The dataset contains four trajectory types: \texttt{random}, \texttt{square}, \texttt{chirp}, and \texttt{melon}. We train on three families (\texttt{random}, \texttt{square}, \texttt{chirp}) using runs 1--3 of each trajectory type for training and run 4 for validation. The held-out \texttt{melon} trajectory (runs 1--3) is reserved entirely for testing, so generalisation is assessed on a trajectory type unseen during training.

Four models are trained on this task, one with each learning algorithm. All four share the single-layer RG-LRU cell of Section~\ref{sec:rglru} with $H = 128$, with three changes appropriate to the nanodrone setting. These include the learnt input projection of \ref{app:rglru} (since $I = 4 \neq H$), the $\tanh$ state-initialisation head of Eq.~\eqref{eq:rglru_state_init} in place of a zero initial state, and a mean squared error (MSE) output term (Eq.~\eqref{eq:rglru_free_energy_reg}) in place of softmax cross-entropy. The model contains 21{,}001 trainable parameters in total. Models are trained on rolling windows of $T = 200$ timesteps (2 seconds overall) with batch size 256 for 50 epochs. Reported values are means and standard deviations over 5 seeds. For full experimental details, see \ref{app:exp_details_nanodrone}.

\paragraph{Results}
\label{sec:nanodrone_results}
Table~\ref{tab:nanodrone_results} reports rollout performance on the held-out \texttt{melon} trajectory. The equivalence observed in Sections~\ref{sec:copy}--\ref{sec:translation} appears here again with tPC-RTRL closely matching BPTT across all three metrics, whereas standard tPC remains close to the spatial BP baseline. Specifically, tPC-RTRL achieves mean position, velocity, and attitude errors of 0.506\,m, 0.808\,m/s, and 0.232\,rad, compared with 0.505\,m, 0.809\,m/s, and 0.230\,rad for BPTT.

This result is important for two reasons. First, it shows that the BPTT/tPC-RTRL relationship is not confined to synthetic long-range benchmarks or discrete-token language tasks, but persists in a realistic continuous setting. Second, it further supports the diagnosis that the key limitation of standard tPC in recurrent settings is temporal credit assignment, and once recurrent influence is tracked through time, PC-based learning recovers BP-like performance even on this challenging system-identification problem.

\begin{table*}[t]
    \centering
    \small
    \begin{tabular}{lccc}
        \toprule
        Method & Mean pos.\ err.\ (m) $\downarrow$ & Mean vel.\ err.\ (m/s) $\downarrow$ & Mean att.\ err.\ (rad) $\downarrow$ \\
        \midrule
        Spatial BP & $0.578 \pm 0.013$ & $0.924 \pm 0.019$ & $0.247 \pm 0.015$ \\
        tPC        & $0.556 \pm 0.015$ & $0.911 \pm 0.020$ & $0.253 \pm 0.006$ \\
        BPTT       & $0.505 \pm 0.015$ & $0.809 \pm 0.037$ & $0.230 \pm 0.006$ \\
        tPC-RTRL   & $0.506 \pm 0.015$ & $0.808 \pm 0.027$ & $0.232 \pm 0.005$ \\
        \bottomrule
    \end{tabular}
    \caption{
        Nanodrone rollout performance on the held-out \texttt{melon} trajectory. Values are mean rollout errors over the full 200-step horizon, reported as mean $\pm$ standard deviation over 5 seeds. Att.\ err.\ denotes rotation-vector error in radians.
    }
    \label{tab:nanodrone_results}
\end{table*}

\subsection{Filtering with tPC-RTRL}
\label{sec:filtering}
A structural feature of tPC-based learning that distinguishes it from purely RTRL-based approaches is that the iterative inference mechanism used during training is also available at deployment time. In a filtering or state-estimation setting, observations of the true state may arrive periodically during deployment, and the model should incorporate these corrections into its internal trajectory. Under tPC-RTRL, the natural way to do this is to clamp the observation as a target in the free energy objective (Eq.~\ref{eq:tpc_free_energy}) and run iterative inference, which revises the internal latent state without any modification to the learning algorithm.

This section presents evidence that this mechanism is useful in practice. The results below are encouraging in the high-correction-frequency regime but become marginal as the correction interval grows, so we identify several concrete directions for future work in Section~\ref{sec:beyond_predictive}.

\paragraph{Task Definition} We take the tPC-RTRL models from Section~\ref{sec:nanodrone_results} and evaluate them on the held-out \texttt{melon} trajectory under a range of test-time correction schedules. At each correction step, the full ground-truth state is revealed. We compare three modes:
\begin{itemize}
    \item \emph{Pure rollout}: no correction, the model evolves autonomously for the full sequence. 
    \item \emph{Amortised}: the latent state is reset directly using the feedforward state-initialisation network. 
    \item \emph{tPC Inference}: the observation is clamped as the target and iterative free energy minimisation is used to update the latent state before rollout continues. 
\end{itemize}
Between corrections, all methods evolve under the same learnt recurrent dynamics. We vary the correction period over $k \in \{10, 25, 50, 100\}$ steps, with an initial observation always taken at $t=0$, and report mean $\pm$ standard deviation over 5 runs.

\begin{figure}[t]
    \centering
    \includegraphics[width=\linewidth]{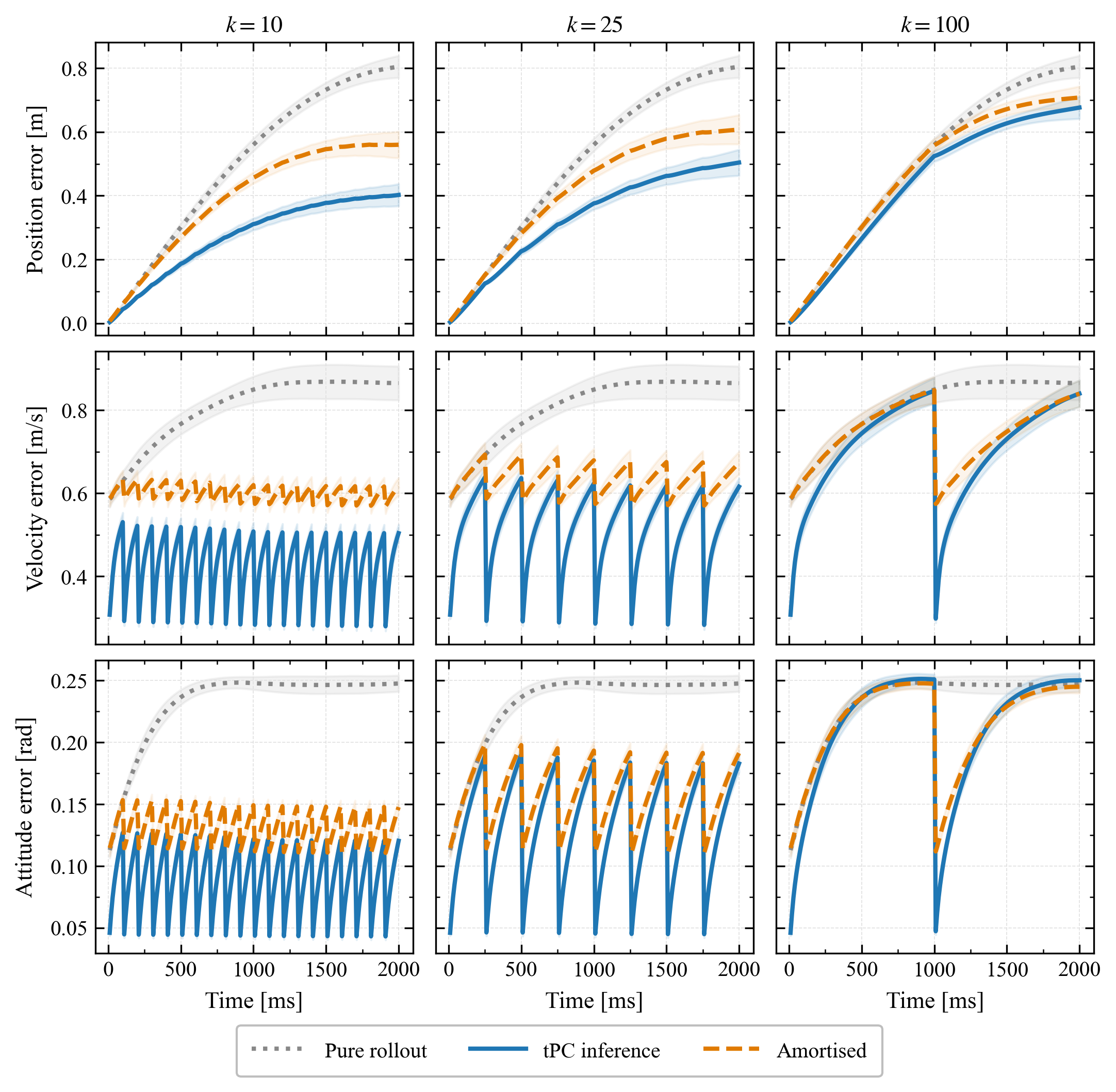}
    \caption{
        Rollout error on the held-out \texttt{melon} trajectory under intermittent test-time correction. Columns show correction periods of $k \in \{10, 25, 100\}$ steps; rows show position, velocity, and attitude error. Grey dotted lines: pure rollout. Dashed lines: amortised correction. Solid lines: tPC inference-based correction. Shaded regions indicate one standard deviation over 5 runs. $k =50$ is omitted for brevity.
    }
    \label{fig:filtering_rollout}
\end{figure}

\paragraph{Results} Table~\ref{tab:filtering_results} and Figure~\ref{fig:filtering_rollout} summarise the outcome. Incorporating intermittent observations at test-time substantially improves over pure open-loop rollout. Table~\ref{tab:filtering_results} shows that with no correction, the model reaches a final position error of 0.805\,$\pm$\,0.034\,m. When $k=10$, tPC inference reduces this to 0.402\,$\pm$\,0.035\,m, compared with 0.560\,$\pm$\,0.042\,m for amortised correction. At the highest correction frequency, inference therefore cuts final position error by approximately half relative to pure rollout, and remains clearly better than a direct reset of the latent state.

The advantage of tPC inference is largest when observations are frequent and decreases as the correction interval grows. At $k=25$, tPC inference outperforms amortised correction (0.504\,vs.\ 0.607\,m final position error) and yields lower rollout-averaged velocity and attitude errors. At $k=50$, the remaining benefit is more modest, tPC inference still improves final position error and slightly improves velocity error, whilst attitude is essentially unchanged. By $k=100$, the two correction schemes are very similar. tPC inference retains a small advantage in final position error (0.676\,vs.\ 0.708\,m) but the differences in rollout-averaged metrics are negligible, and inference is marginally worse than amortised correction in final attitude error (bottom right of Figure~\ref{fig:filtering_rollout}).

The fact that tPC inference is consistently better than the amortised reset at high correction frequencies, with no change to the learning algorithm or architecture, is encouraging evidence that the filtering interpretation of tPC-RTRL may have some practical utility. However, the narrowing advantage at longer correction intervals suggests further investigation is needed. We hypothesise two contributing causes. First, the nanodrone benchmark \cite{Busetto_2026} has aggressive flight characteristics, and even with intermittent ground-truth corrections the drone state is difficult to track over relatively short timescales. This suggests that even a perfect correction mechanism is likely to degrade quickly. Second, the model has been trained in the predictive-rollout regime. Therefore, during training, it experiences only regions of latent space reachable by the feedforward transition prior, so an inferred state applied at test-time may fall outside the region of the latent-state manifold visited during training. This means that the test-time distribution of latent states differs from that seen during training, and could possibly cause some degradation in subsequent rollout performance, providing a potential explanation for the degradation observed in the $k=100$ attitude error (Figure \ref{fig:filtering_rollout}, bottom right). 

\begin{table*}[t]
    \centering
    \small
    \setlength{\tabcolsep}{5pt}
\begin{tabular}{llccc}
    \toprule
    $k$ & Method & Final pos.\ err.\ (m) $\downarrow$ & Mean vel.\ err.\ (m/s) $\downarrow$ & Mean att.\ err.\ (rad) $\downarrow$ \\
    \midrule
    --- & Pure rollout
        & $0.805 \pm 0.034$
        & $0.808 \pm 0.027$
        & $0.233 \pm 0.005$ \\
    \midrule
    \multirow{2}{*}{10}
        & Amortised
        & $0.560 \pm 0.042$
        & $0.601 \pm 0.017$
        & $0.132 \pm 0.005$ \\
        & tPC Inference
        & $0.402 \pm 0.035$
        & $0.427 \pm 0.019$
        & $0.089 \pm 0.004$ \\
    \midrule
    \multirow{2}{*}{25}
        & Amortised
        & $0.607 \pm 0.046$
        & $0.633 \pm 0.022$
        & $0.158 \pm 0.006$ \\
        & tPC Inference
        & $0.504 \pm 0.041$
        & $0.519 \pm 0.016$
        & $0.132 \pm 0.004$ \\
    \midrule
    \multirow{2}{*}{50}
        & Amortised
        & $0.652 \pm 0.042$
        & $0.678 \pm 0.023$
        & $0.187 \pm 0.006$ \\
        & tPC Inference
        & $0.586 \pm 0.041$
        & $0.607 \pm 0.016$
        & $0.174 \pm 0.006$ \\
    \midrule
    \multirow{2}{*}{100}
        & Amortised
        & $0.708 \pm 0.035$
        & $0.744 \pm 0.021$
        & $0.216 \pm 0.005$ \\
        & tPC Inference
        & $0.676 \pm 0.035$
        & $0.705 \pm 0.018$
        & $0.211 \pm 0.006$ \\
    \bottomrule
\end{tabular}
    \caption{
        Test-time filtering performance on the held-out \texttt{melon} trajectory using the tPC-RTRL trained models. $k$ denotes the number of rollout steps between corrections. Amortised correction resets the latent state using the feedforward state-initialisation network; tPC inference correction updates the latent state via iterative free energy minimisation. Reported as mean $\pm$ standard deviation over 5 runs.
    }
    \label{tab:filtering_results}
\end{table*}

\section{Discussion}
\label{sec:discussion}

\subsection{Locality in space and time}
\label{sec:locality}
The central conceptual contribution of this paper is that tPC-RTRL combines spatially local PC error signals with temporally local RTRL-style influence tracking.

Spatial locality is the standard appeal of PC where the global backward pass of BP is replaced by local prediction-error interactions at each layer. Each layer updates its parameters using only quantities available at its own inputs and outputs, without requiring a globally coordinated reverse traversal of the forward graph. RTRL alone does not provide this. To obtain error signals at the recurrent state, a plain RTRL system must still backpropagate through the readout network, which requires the same phase-separated, globally coordinated backward pass that makes BPTT unattractive for neuromorphic hardware.

The temporal locality of RTRL is one of its most appealing properties. The full stored activation trajectory required by BPTT is replaced by an online influence matrix maintained forward in time. The memory cost becomes independent of sequence length, and no explicit unrolling is required. tPC alone does not provide this. We demonstrate in Sections~\ref{sec:copy}--\ref{sec:nanodrone}, that standard tPC's one-step recurrent update fails to propagate credit over long horizons and behaves like spatial BP.

Because the two mechanisms address orthogonal parts of the problem, their composition yields a credit-assignment mechanism that inherits the relevant locality properties of both components. Proposition~\ref{prop:tpcrtrl_bptt}, in Section~\ref{sec:bptt_relation}, states the idealised consequence, under explicit assumptions, the composition recovers BPTT gradients exactly, and the experiments in Sections~\ref{sec:copy}--\ref{sec:nanodrone} provide empirical support that the algorithm survives relaxation of those assumptions in practice. A secondary structural benefit arises in filtering settings. As tPC already contains iterative inference as part of its learning machinery, the same mechanism can be reused at deployment time to incorporate online state corrections, as explored in Section~\ref{sec:filtering}. A pure RTRL system would require a separate mechanism for this, however the tPC-RTRL formulation readily permits this capability without further modification to the algorithm.

\subsection{Memory implications}
\label{sec:memory}
\begin{figure}[t]
    \centering
    \includegraphics[width=\linewidth]{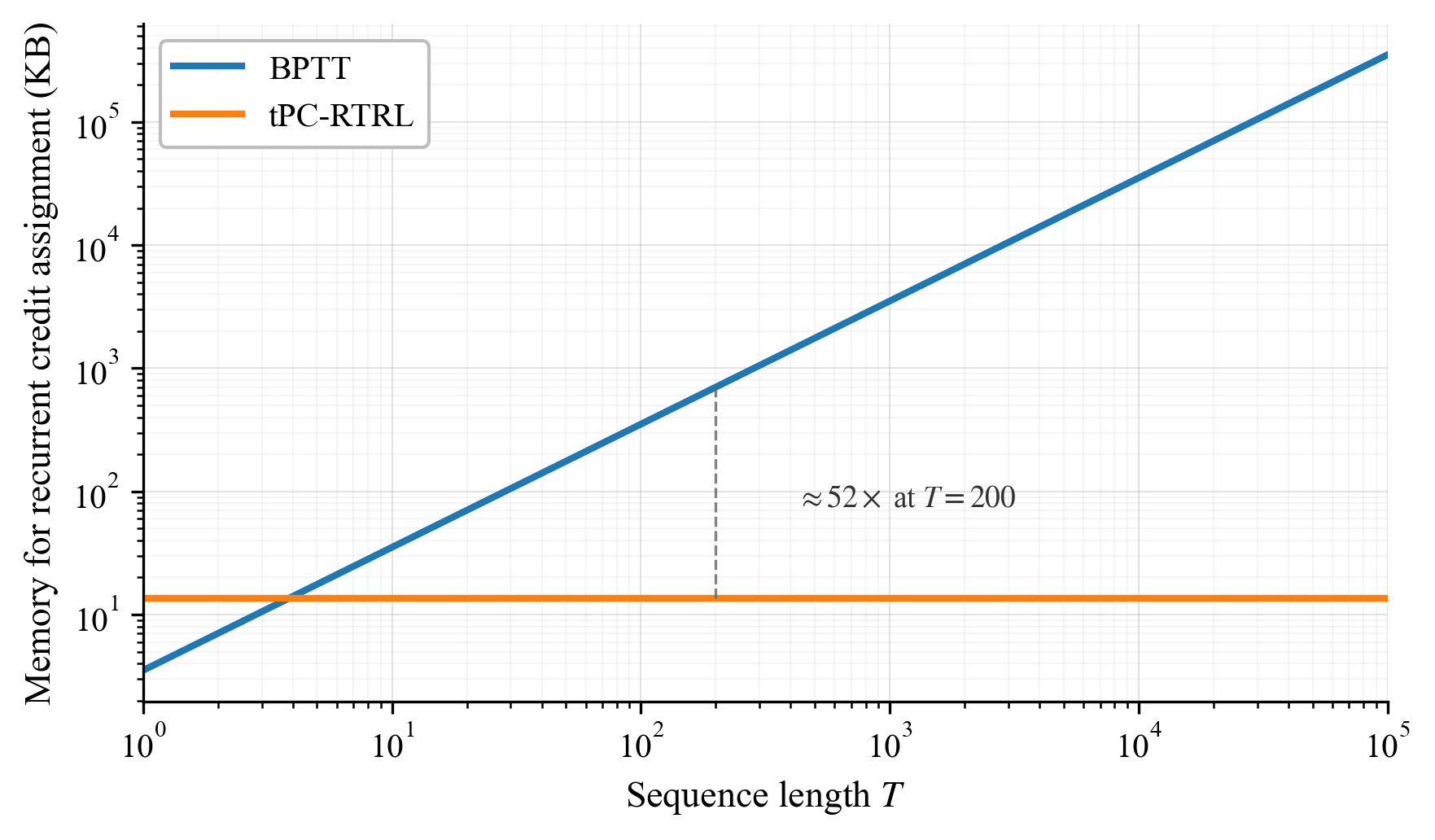}
    \caption{
    Memory scaling for recurrent credit assignment in the nanodrone RG-LRU model ($H=128$, $I=4$). BPTT memory grows as $TI+7TH$, while tPC-RTRL remains constant at 3{,}456 stored values (approximately 13.5\,KB). At $T=200$, BPTT requires approximately 703\,KB, giving a $52\times$ reduction for tPC-RTRL.
    }
    \label{fig:recurrent_memory_scaling}
\end{figure}

The temporal locality properties translate into a concrete memory implication (Figure~\ref{fig:recurrent_memory_scaling}). For the recurrent part of the nanodrone RG-LRU model used in Sections~\ref{sec:nanodrone}--\ref{sec:filtering}, BPTT retains the time-indexed activations needed to backpropagate through the recurrent trajectory. Excluding the readout pathway and focusing only on the state-transition part of the model, this consists of the input sequence $\xx_L^{(1:T)} \in \mathbb{R}^{T \times I}$ and seven recurrent tensors in $\mathbb{R}^{T \times H}$: the projected input $p^{(t)}$, the two gates $g_z^{(t)}$ and $g_a^{(t)}$, the decay quantities $\log a^{(t)}$, $a^{(t)}$, and $\gamma^{(t)}$, and the recurrent state sequence $\bar{\xx}^{(1:T)}$. This gives
\[
T I + 7 T H
= 200 \cdot 4 + 7 \cdot 200 \cdot 128
= 180{,}000
\]
stored values, or approximately 703\,KB at float32. By contrast, tPC-RTRL, on the element-wise RG-LRU,  stores only the current recurrent state $\bar{\xx}^{(t)}$ together with the influence traces, $\mathbf{M}_{\theta}^{(t)}$, maintained online by the recurrent tracker. In this implementation those traces correspond to parameters $\Lambda$, $W_a$, $b_a$, $W_z$, $b_z$, $W_{\mathrm{in}}$, $b_{\mathrm{in}}$, $W_{x_0}$, and $b_{x_0}$, for a total of 3{,}328 influence values. Including the current 128-dimensional recurrent state gives 3{,}456 stored values in total, or approximately 13.5\,KB, independent of sequence length.

At $T = 200$, this is about a 52$\times$ reduction in the storage required for temporal credit assignment relative to BPTT. Furthermore, the BPTT requirement grows with $T$, at $T = 1{,}000$ it rises to approximately 3{,}516\,KB, and at $T = 10{,}000$ to approximately 35{,}156\,KB, whereas tPC-RTRL remains at 13.5\,KB throughout. For streaming or long-horizon deployment scenarios, this sequence-length independence becomes significant.

This comparison concerns storage for recurrent credit assignment only; it does not include compute time, inference iterations, optimiser state, or non-recurrent activations.

\subsection{Equivalence with BPTT}
\label{sec:near_equivalence}
Proposition~\ref{prop:tpcrtrl_bptt} establishes that tPC-RTRL coincides with BPTT exactly under the fixed-prediction assumption \cite{DBLP:journals/corr/abs-2006-04182} and converged inference. As discussed in Section~\ref{sec:bptt_relation}, our experiments do not enforce this regime, yet tPC-RTRL tracks BPTT closely across four tasks. For each experiment, we hand-tuned the number of inference iterations and the inference momentum to maximise the cosine similarity between the tPC-RTRL and BPTT gradients on a single batch of untrained-model data. This served as a credible proxy for a full hyperparameter sweep over the inference dynamics, and was used to effectively select the inference parameters throughout Section~\ref{sec:results}. At these points, the cosine similarity between PC and BP gradient vectors on the recurrent parameters remains relatively close to 1 throughout training (see \ref{app:cosine_similarity} for the measured similarity over training on WikiText-103), so the direction of the tPC-RTRL update is largely preserved relative to BPTT.

The residual differences appear mostly in magnitude rather than direction. Because at the start of inference, all the energy is concentrated at the network output, the per-layer norm of the PC gradient typically decreases from output to input, and the resulting tPC-RTRL update inherits a similar layer-wise magnitude profile. Adam's per-parameter second-moment normalisation \citep{kingma2017adammethodstochasticoptimization} absorbs such magnitude differences between the otherwise well-aligned gradient vectors. This combination, with high-cosine-similarity directions from the tuned inference dynamics, and adaptive per-parameter scaling from Adam, appears sufficient to minimise any resulting optimisation trajectory difference between BPTT and tPC-RTRL on the metrics tested.

What tPC-RTRL achieves in practice is therefore not strictly equivalent to BPTT. Proposition~\ref{prop:tpcrtrl_bptt} establishes that the BPTT gradient lies within the family of updates tPC-RTRL can produce under tightly controlled assumptions. The empirical results then show that, at a tuned but practical operating point, the updates tPC-RTRL produces are aligned closely enough in direction, and Adam tolerant enough of the magnitude residuals, that BPTT-equivalent optimisation behaviour emerges without enforcing the proof's assumptions. This framing is consistent with PC's broader role in the literature, where its value in certain settings is reported to lie outside the strict-equivalence regime \citep{song_inferring_2024,mali2024tightstabilityconvergencerobustness}. Our results do not make any claim related to this, they simply report that, within the variance of our experiments, no measurable difference between BPTT and tPC-RTRL emerges on the tasks tested, and that the two methods track each other closely across training. Exploring conditions under which material differences between the two methods emerge, such as on significantly deeper models, and if tPC-RTRL carries the advantages of PC-based learning beyond the BPTT-equivalence regime, are identified as important directions for future work.

\subsection{Scope of the hardware argument}
\label{sec:hardware_scope}
We have motivated tPC-RTRL partly through its suitability for resource-constrained deployment. We frame this as a structural compatibility argument rather than an experimentally established systems advantage as we do not report energy measurements, FLOP costs, or deployment results on neuromorphic hardware. Translating the structural properties established above into measured energy or latency advantages on real hardware involves many additional factors, including memory bandwidth, inter-core communication costs, and the overhead of the influence-matrix update itself. Characterising the full hardware profile of tPC-RTRL on a platform such as Loihi 2 \cite{Orchard2021Loihi2} or SpiNNaker 2 \cite{Gonzalez2024SpiNNaker2} is a concrete and important direction for future work.

\subsection{Architectural extensions}
\label{sec:architectural_extensions}
All experiments in this work use a single-layer RG-LRU recurrent module. This is a deliberate choice consistent with the edge-deployment target, but it is worth discussing how the approach extends to deeper recurrent stacks. For an $L$-layer stack, each hidden layer depends on the parameters of all layers below, and a naive extension of exact RTRL must maintain $L(L+1)/2$ influence matrices \cite{irie2024exploringpromiselimitsrealtime}. Element-wise recurrent modules help limit the size of each individual influence matrix, but the inter-layer dependency structure means that higher layers cannot be computed from the local perspective of a single layer alone. Preserving the intra-layer locality that makes tPC-RTRL attractive for neuromorphic hardware therefore requires an approximation.

A promising route is the layer-local RTRL approximation of \citet{zucchet2023onlinelearninglongrangedependencies}, in which each recurrent layer maintains its own influence matrix and cross-layer influence is ignored. The authors show that this approximation retains much of the expressivity gain of deeper networks in conventional recurrent architectures. This approximation composes naturally with tPC-RTRL as each layer still receives a local prediction-error signal from PC, and each layer's parameter update remains computable from quantities local to that layer. A multi-layer tPC-RTRL system would therefore be local at the layer level in both the spatial and temporal axes, at the cost of exactness. Characterising the resulting gradient gap, and studying at what depths it becomes practically consequential, is a natural follow-up. Beyond the RG-LRU, tPC-RTRL could also be combined with other element-wise recurrent units \cite{irie2024exploringpromiselimitsrealtime, orvieto2023resurrectingrecurrentneuralnetworks} or with sparse or factorised approximations of RTRL \cite{marschall2019unifiedframeworkonlinelearning}. Progress on approximate or structured RTRL methods is therefore complementary to the tPC-RTRL framework proposed here.

\subsection{Beyond predictive-rollout}
\label{sec:beyond_predictive}
We adopt the predictive-rollout regime throughout, propagating $\bar{\xx}^{(t)} = \mu_x^{(t)}$ rather than the inferred state. This ensures consistency between training and test-time dynamics, incentivises the model to build good rollout performance rather than relying on corrections, and keeps the RTRL recursion exact. However, during learning, iterative inference serves only as a means to obtain local parameter gradients, and does not revise the latent trajectory the model follows. The filtering results of Section~\ref{sec:filtering} begin to suggest this may have implications. When an inferred-state correction is applied at test-time to a model trained under predictive-rollout, the corrected state may fall outside the region of latent space the learnt dynamics have experienced during training. In our filtering experiments this effect was small, visible only as a slight increase in final attitude error at the longest correction interval ($k = 100$, Table~\ref{tab:filtering_results}). However, it may present a more significant issue in other settings, perhaps in scenarios where open-loop rollout performance degrades less quickly, and intermittent corrections have a longer lasting influence on the trajectory. Characterising where the solutions found by iterative inference sit in the latent space relative to the feedforward rollout trajectory, to what degree the recurrent prior anchors the inferred state to familiar regions of this space, and quantifying the downstream consequences for rollout performance, are important directions for future work.

A related question is whether the training regime itself should adapt to inferred-state dynamics. \ref{app:inferred-propagation} examines the inferred-propagation regime, in which $\hat{\xx}^{(t)}$ is propagated throughout training under the approximate influence recursion of Eq.~\eqref{eq:tpcrtrl_influence}. On the copy task, the resulting gradient bias produces a modest degradation but still improves on standard tPC (\ref{app:inferred-gradient-bias}); on the nanodrone benchmark, training loss drops well below the predictive-rollout baseline whilst open-loop validation rollout substantially worsens. This is because inference corrections continuously repair the latent trajectory and the recurrent dynamics are no longer pressured to learn accurate autonomous rollouts (\ref{app:inferred-nanodrone}). These small experiments suggest some areas for future work. First, quantifying how the approximation-induced gradient bias scales with distance between the inferred state and the predictive-rollout trajectory, and how this interacts with the training dynamics, is a necessary step toward understanding the conditions under which the induced bias is benign or harmful. Second, exploring hybrid regimes, analogous to scheduled sampling \cite{bengio2015scheduledsamplingsequenceprediction}, that begin under predictive-rollout before gradually introducing inferred-state propagation, might allow the recurrent dynamics to first acquire the autonomous-rollout competence that predictive-rollout incentivises, before then gradually exposing the model to the inferred-state distribution and encouraging it to learn dynamics that are robust to this shift. We did not pursue these further here because the filtering results of Section~\ref{sec:filtering} suggest much of the practical benefit of iterative inference can be obtained at test-time without altering the training regime, so the immediate incentive to redesign training was limited; whether this changes for longer horizons or different rollout-degradation profiles remains open and we leave these as promising directions for future work.

\section{Conclusion}
This work introduced tPC-RTRL, an algorithm for recurrent neural networks that combines the local, parallelisable credit assignment of Temporal Predictive Coding (tPC) with an online influence matrix that tracks the recursive contribution of recurrent parameters across time. We showed that, under explicit assumptions, tPC-RTRL recovers the exact gradients computed by backpropagation-through-time (BPTT). Empirically, across four tasks including a controlled copy task, byte-level language modelling on WikiText-103, English--French translation on CCMatrix, and a realistic nanodrone system-identification benchmark, the observations are consistent. tPC-RTRL matches BPTT very closely, and standard tPC tracks one-step truncated BPTT. Filtering experiments on a nanodrone benchmark further show that the iterative inference mechanism used during training can also be used at test-time to incorporate intermittent state observations.

The broader conceptual claim is that Predictive Coding (PC) and Real-Time Recurrent Learning (RTRL) address complementary aspects of recurrent credit assignment. PC supplies local spatial error signals, while RTRL supplies online temporal influence tracking. Together, they yield a recurrent learning rule with locality properties that neither component provides in isolation. These properties align with the communication, memory, and synchronisation constraints that motivate neuromorphic and edge-hardware implementations. Their combination therefore extends tPC beyond one-step temporal credit assignment without discarding the inference-based structure that makes PC attractive.

This combination is relevant to continuously operating systems in which data arrive sequentially, memory is constrained, and intermittent observations may be available to correct an evolving internal state, including adaptive robotics, autonomous aerial vehicles, industrial monitoring, and wearable or edge-based sensing. In such settings, RTRL avoids BPTT's need to retain and reverse-traverse an unrolled state trajectory, and the PC component provides a mechanism for incorporating state observations at deployment time using the same iterative inference dynamics used during training. For the single-layer nanodrone model, this corresponds to an approximate 52$\times$ reduction in the storage required specifically for temporal credit assignment relative to BPTT at a 200-step horizon, with this requirement independent of sequence length. More generally, tPC-RTRL suggests that long-range learning and observation-conditioned state correction can be brought together within a common framework.

Several directions for future work are open. First, extending tPC-RTRL to deeper recurrent stacks via the layer-local RTRL approximation of \citet{zucchet2023onlinelearninglongrangedependencies}. Second, a further investigation of the inferred-state propagation regime, including quantifying the approximation-induced gradient bias and exploring hybrid training regimes that gradually introduce inferred-state propagation. Third, whilst the structural arguments for compatibility with neuromorphic and edge hardware are concrete, demonstrating measured energy or latency advantages on real hardware, for example on Loihi 2 \cite{Orchard2021Loihi2} or SpiNNaker 2 \cite{Gonzalez2024SpiNNaker2}, remains essential future work. Addressing these questions represents the natural next step toward energy-efficient on-device training of recurrent sequence models.

\section{Funding}
This research was supported by the UK Government through the EPSRC Edgy Organism project (EP/Y030133/1).

\bibliographystyle{elsarticle-num-names}
\bibliography{tpc}

\clearpage
\appendix

\section*{Appendix}
This Appendix provides additional technical detail supporting the methods and empirical results presented in the main paper. The main text is intended to be self-contained; the material collected here is provided to support reproducibility and to clarify the derivations, implementation choices, and experimental procedures underlying Temporal Predictive Coding with Real-Time Recurrent Learning (tPC-RTRL).

\section{Derivation of the PC free energy and local update rules}
\label{app:pc_derivation}

This appendix gives the full derivation summarised in
Section~\ref{sec:pc_background} including the variational free energy
formulation of PC, its reduction to a sum of local squared
prediction errors under Gaussian conditionals and the mean-field
approximation, and the resulting layer-local inference and learning
rules.

PC can be viewed as performing approximate Bayesian inference in a
hierarchical generative model. We consider a discriminative setting,
in which the input is treated as the top-level latent variable,
$\xx_L$, whilst the observed target is denoted $\yy$. The remaining
variables $\xx_{1:L-1}$ are latent and must be inferred. Writing the
conditional generative model as
\begin{equation}
p(\yy, \xx_{1:L-1} \mid \xx_L;  \theta),
\end{equation}
exact inference would require the posterior $p(\xx_{1:L-1} \mid \yy,
\xx_L; \theta)$, which is generally intractable. PC introduces a
variational posterior $q(\xx_{1:L-1})$ and optimises the variational
free energy
\begin{equation}
\F[q, \theta]
=
\mathbb{E}_{q}
\left[
\log q
-
\log p(\yy, \xx_{1:L-1} \mid \xx_L; \theta)
\right].
\end{equation}
Equivalently, $\F$ is the negative evidence lower bound
\begin{align}
\F[q, \theta]
=
&\mathrm{KL}\!\left[
q
\;\middle\|\;
p(\xx_{1:L-1} \mid \yy, \xx_L; \theta)
\right]\notag\\
&-
\log p(\yy \mid \xx_L; \theta).
\end{align}
Minimising $\F$ with respect to $q$ is therefore equivalent to
minimising the Kullback–Leibler (KL) divergence to the exact posterior, since the
evidence term is constant with respect to $q$.

To obtain the standard PC objective, we specify a layer-wise
factorisation of the generative model with Gaussian conditionals,
\begin{equation}
    p(\xx_l \mid \xx_{l+1}; \theta_l)
    =
    \mathcal{N}(\xx_l; g_l(\xx_{l+1}; \theta_l), \Sigma_l),
\end{equation}
for $l=0,\dots,L-1,$ with $\xx_0 \equiv \yy$. Assuming that $q$
factorises under a mean-field approximation as
\begin{equation}
    q(\xx_{1:L-1}) = \prod_{l=1}^{L-1} q_l(\xx_l),
\end{equation}
and that each factor is modelled either as a Gaussian with
sufficiently small variance or as a Dirac delta centred on the
current state estimate, $\F$ reduces to a sum of precision-weighted
local prediction errors:
\begin{align}
&\F
=
\frac{1}{2}
\sum_{l=0}^{L-1}
\epsilon_l^\top \Pi_l \epsilon_l,
\\&\epsilon_l
=
\xx_l - g_l(\xx_{l+1};\theta_l),
\end{align}
with $\Pi_l = \Sigma_l^{-1}$. It is common for the precision terms
to be fixed to the identity, in which case they can be omitted, and
we do so throughout this work. The free energy objective therefore
simplifies to
\begin{equation}
\F
=
\frac{1}{2}
\sum_{l=0}^{L-1}
\epsilon_l^\top \epsilon_l.
\end{equation}

PC is typically implemented using an Expectation-Maximisation (EM) procedure
\cite{Whittington2017AnAO}. First, with weights fixed, the latent
states $\xx_{1:L-1}$ are iteratively updated to reduce $\F$ (an
E-step). Then, with the states held fixed at their inferred values,
the model parameters are updated by descending the same objective
with respect to $\theta$ (an M-step). Inference and learning therefore
correspond to alternating optimisation over variational parameters
and model parameters of a single free energy functional.

Examining $\F$ makes clear why PC gives rise to layer-local rules for
inference and learning. Inference corresponds to gradient descent on
$\F$ with respect to the latent activities $\xx_{1:L-1}$. For an
internal layer $l \in \{1,\dots,L-1\}$,
\begin{equation}
    \Delta \xx_l
    =
    -\alpha \frac{\partial \F}{\partial \xx_l}
    =
    \alpha
    \left(
        \left(
        \frac{\partial g_{l-1}(\xx_l;\theta_{l-1})}{\partial \xx_l}
        \right)^\top
        \epsilon_{l-1}
        -
        \epsilon_l
    \right),
\end{equation}
so the update to each state depends only on the prediction error at
its own layer and the layer directly below. Similarly, each layer
updates its weights using only locally available quantities:
\begin{equation}
    \Delta W_l
    =
    -\eta \frac{\partial \F}{\partial \theta_l}
    =
    \eta
    \left(
        \frac{\partial g_l(\xx_{l+1};\theta_l)}{\partial \theta_l}
    \right)^\top
    \epsilon_l.
\end{equation}

When $g_l$ takes a particular form we recover a Hebbian-like update
rule. Consider a pre-activation parameterisation of the downward
prediction,
\begin{equation}
    g_l(\xx_{l+1};\theta_l) =  \theta_l\sigma(\xx_{l+1}).
\end{equation}
In this case, the parameter update remains local and Hebbian-like:
\begin{equation}
    \Delta \theta_{l,ij}
    \propto
    \sigma(\xx_{l+1})_j  \epsilon_{l,i}.
\end{equation}
This can be interpreted as a postsynaptic prediction-error term
modulating nonlinear presynaptic activity.

\section{Algorithmic details for tPC-RTRL}
\label{app:tpcrtrl_algorithm}
All parameter updates in this work follow the forward-update
convention of \citet{qi2025training}. This method is known to stabilise
training in deeper PC networks~\citep{qi2025training}. For clarity, and compatibility with more general PC works, all definitions are written in the standard free energy
form; the forward-update convention does not change the inference procedure and determines only how the
parameter derivatives are realised during the M-step. In practice this convention means that during the weight update, rather than evaluating the free energy error terms at the current layer-wise errors, given by the mismatch between the inferred state and its prediction, we evaluate them at the mismatch between the initial feedforward prediction and the inferred state. As an example with the readout error term from the network defined in Figure \ref{fig:tpcrtrl_diagram}, the standard error term would be
\begin{equation}
\epsilon_y^{(t)} = \yy^{(t)} - g(\hat{\xx}^{(t)}; \theta_{\mathrm{out}}),
\end{equation}
where $\hat{\xx}^{(t)}$ is the inferred latent state. Under the forward-update convention, this becomes
\begin{equation}
\epsilon_y^{(t)} = \yy^{(t)} - g(\xx^{(t, 0)}; \theta_{\mathrm{out}}),
\end{equation}
where $\xx^{(t, 0)}$ is the value of $\xx^{(t)}$ at the start of inference, before any updates to the latent state have been applied. \citet{qi2025training} show that this convention keeps the parameter updates anchored to the feedforward trajectory that the model follows during test-time rollout, and therefore improves performance. In practice, the additional cost of evaluating the free energy at the feedforward prediction rather than the inferred state is very small, as it only requires storing the feedforward prediction at the start of inference, and discarding it once calculating the parameter gradients.

\section{RG-LRU implementation details}
\label{app:rglru}
This appendix further details the 
Real-Gated Linear Recurrent Unit (RG-LRU) \cite{de2024griffinmixinggatedlinear} introduced in Section~\ref{sec:rglru}, this includes the explicit gradients and momentum-accelerated dynamics used during PC inference, the regression-output variant of the per-timestep free energy used on the nanodrone benchmark, and the optional input-projection and learnt state-initialisation heads. Notation follows Section~\ref{sec:rglru} throughout where $\xx_L^{(t)} \in \mathbb{R}^{I}$ is the cell input, $\bar{\xx}^{(t-1)} \in \mathbb{R}^{H}$ is the propagated recurrent state, $\mu_x^{(t)}$ is the recurrent prediction, $\xx^{(t)}$ is the inferred recurrent latent, and $\mathbf{o}^{(t)}$ is the inferred readout latent.

\paragraph{Input projection} When the input dimensionality $I$ differs from the recurrent state dimensionality $H$, the input that enters the gated-input pathway of Eq.~\eqref{eq:rglru_transition} is projected to $\mathbb{R}^H$:
\begin{equation}
p^{(t)} = W_{\mathrm{in}}\, \xx_L^{(t)} + b_{\mathrm{in}}, \qquad W_{\mathrm{in}} \in \mathbb{R}^{H \times I},\ b_{\mathrm{in}} \in \mathbb{R}^{H},
\end{equation}
and the recurrent update becomes
\begin{equation}
\mu_x^{(t)} = a^{(t)} \odot \bar{\xx}^{(t-1)} + \gamma^{(t)} \odot \bigl(g_z^{(t)} \odot p^{(t)}\bigr).
\label{eq:rglru_transition_proj}
\end{equation}
The models in Sections \ref{sec:wikitext} and \ref{sec:translation} have $I = H$ and use $W_{\mathrm{in}} = \mathbb{I}$, $b_{\mathrm{in}} = 0$, in which case the projection is no longer a learnable parameter. The nanodrone model (Sections \ref{sec:nanodrone} and \ref{sec:filtering}) uses $I = 4$, $H = 128$ and learns $\{W_{\mathrm{in}}, b_{\mathrm{in}}\}$ jointly with $\theta_{\mathrm{rec}}$. In what follows we treat $p^{(t)}$ uniformly, with $p^{(t)} = \xx_L^{(t)}$ understood in the un-projected case.

\paragraph{learnt state initialisation} For the nanodrone model (Sections \ref{sec:nanodrone} and \ref{sec:filtering}) the recurrent state at the start of each sequence is produced from the observed body-frame state $s_0 \in \mathbb{R}^9$ by a feedforward head
\begin{equation}
\bar{\xx}^{(0)} = \tanh(W_{x_0}\, s_0 + b_{x_0}),
\label{eq:rglru_state_init}
\end{equation}
with $W_{x_0} \in \mathbb{R}^{H \times 9}$, $b_{x_0} \in \mathbb{R}^H$. The other instances of the RG-LRU model use a fixed zero initial state and do not require this component.

\subsection{PC inference}
\label{app:rglru_inference}

The full forward pathway at each timestep consists of the recurrent prediction of Eq.~\eqref{eq:rglru_transition} followed by a ReLU readout and a linear output head. For the copy task (Section \ref{sec:copy}), WikiText-103 (Section \ref{sec:wikitext}), and translation (Section \ref{sec:translation}) models the output head is a softmax classifier with cross-entropy loss, as written in Eq.~\eqref{eq:rglru_free_energy}. For the nanodrone model (Section \ref{sec:nanodrone}) the output head is linear and the output term is the squared error
\begin{equation}
\F^{(t)}_{\mathrm{out}} = \frac{1}{2C}\, \| W_l\, \mathbf{o}^{(t)} + b_l - \yy^{(t)} \|_2^2,
\label{eq:rglru_free_energy_reg}
\end{equation}
which replaces the cross-entropy term in Eq.~\eqref{eq:rglru_free_energy}. In both cases the recurrent and readout terms are unchanged. The normalising constants $1/H, 1/R, 1/C$ arise from treating each latent as a per-coordinate Gaussian with isotropic precision and absorbing the dimensionality into the coefficient.

Inference is initialised at the feedforward values $\xx^{(t)} \leftarrow \mu_x^{(t)}$ and $\mathbf{o}^{(t)} \leftarrow \phi(W_r \mu_x^{(t)} + b_r)$, and minimises $\F^{(t)}$ by a fixed number of momentum-accelerated gradient-descent steps with step size $\alpha$ and momentum coefficient $\beta$. Writing the local prediction errors as
\begin{align}
\epsilon_x^{(t)} &= \xx^{(t)} - \mu_x^{(t)}, \\
\epsilon_o^{(t)} &= \mathbf{o}^{(t)} - \phi(W_r \xx^{(t)} + b_r),
\end{align}
the gradient of $\F^{(t)}$ with respect to each latent is
\begin{align}
\frac{\partial \F^{(t)}}{\partial \xx^{(t)}}
&=
\frac{1}{H}\,\epsilon_x^{(t)}
-
\frac{1}{R}\,W_r^\top\!\bigl(\epsilon_o^{(t)} \odot m_r^{(t)}\bigr),
\label{eq:rglru_grad_x}
\\
\frac{\partial \F^{(t)}}{\partial \mathbf{o}^{(t)}}
&=
\frac{1}{R}\,\epsilon_o^{(t)}
+
\frac{1}{C}\,W_l^\top\!\bigl(\hat{\yy}^{(t)} - \tilde{\yy}^{(t)}\bigr),
\label{eq:rglru_grad_o}
\end{align}
where $m_r^{(t)}$ is the ReLU mask from the readout. In the classification case $\hat{\yy}^{(t)} = \mathrm{softmax}(W_l \mathbf{o}^{(t)} + b_l)$ and $\tilde{\yy}^{(t)}$ is the (optionally label-smoothed) one-hot target; in the regression case $\hat{\yy}^{(t)} = W_l \mathbf{o}^{(t)} + b_l$ and $\tilde{\yy}^{(t)} = \yy^{(t)}$. Each inference step applies
\begin{align}
v_x &\leftarrow \beta\,v_x + \partial \F^{(t)} / \partial \xx^{(t)},
\quad
\xx^{(t)} \leftarrow \xx^{(t)} - \alpha\,v_x,
\\
v_o &\leftarrow \beta\,v_o + \partial \F^{(t)} / \partial \mathbf{o}^{(t)},
\quad
\mathbf{o}^{(t)} \leftarrow \mathbf{o}^{(t)} - \alpha\,v_o.
\end{align}
The number of inference steps and values of $\alpha$, $\beta$ are task-specific and reported in \ref{app:exp_details}.

Once inference has converged to $(\hat{\xx}^{(t)}, \hat{\mathbf{o}}^{(t)})$, the corresponding errors $\hat{\epsilon}_x^{(t)}, \hat{\epsilon}_o^{(t)}$ supply the local signals used to update the readout and output-head parameters via the standard tPC rule $\Delta W = -\eta\,\partial \F^{(t)} / \partial W$. The recurrent state error $\hat{\epsilon}_x^{(t)}$ is additionally the quantity fed into the RTRL tracker of \ref{app:rglru_rtrl} to update the recurrent parameters of the RG-LRU cell.

\subsection{Real-time recurrent learning for the RG-LRU}
\label{app:rglru_rtrl}

The recurrent parameters of the RG-LRU cell are $\theta_{\mathrm{rec}} = \{\Lambda, W_a, b_a, W_z, b_z\}$, together with the input-projection parameters $\{W_{\mathrm{in}}, b_{\mathrm{in}}\}$ when the projection is used and the state-initialisation parameters $\{W_{x_0}, b_{x_0}\}$ when a learnt initialisation is used. For each such parameter we maintain an influence tensor $\mathbf{M}_{\theta}^{(t)} = \partial \bar{\xx}^{(t)} / \partial \theta$, updated online after every timestep.

\paragraph{General recursion} Because the cell update of Eq.~\eqref{eq:rglru_transition} is element-wise in the hidden dimension, every influence tensor decays multiplicatively by $a^{(t)}$ along its hidden axis. The exact RTRL recursion of Eq.~\eqref{eq:rglru_rtrl}, restated here for any $\theta \in \theta_{\mathrm{rec}}$, is
\begin{equation}
\mathbf{M}_{\theta}^{(t)}
=
a^{(t)} \odot \mathbf{M}_{\theta}^{(t-1)}
+
\frac{\partial \mu_x^{(t)}}{\partial \theta^{(t)}}.
\label{eq:rglru_rtrl_general}
\end{equation}
The element-wise structure of the recurrence means each hidden unit depends only on its own row of $W_a, W_z, W_{\mathrm{in}}$ and its own entry of $\Lambda, b_a, b_z, b_{\mathrm{in}}$, so no cross-unit influences need to be stored. The total recurrent storage is therefore $O(H + HI)$ per example, as summarised in Section~\ref{sec:memory}.

The immediate term $\partial \mu_x^{(t)} / \partial \theta^{(t)}$ is obtained directly by differentiating Eq.~\eqref{eq:rglru_transition} (or Eq.~\eqref{eq:rglru_transition_proj} when the input projection is enabled); when the projection is omitted, $W_{\mathrm{in}}, b_{\mathrm{in}}$ drop out of $\theta_{\mathrm{rec}}$ and from the RTRL tracker.

\paragraph{State-initialisation head} For the nanodrone model the initial recurrent state $\bar{\xx}^{(0)}$ is produced by Eq.~\eqref{eq:rglru_state_init}. Because $\{W_{x_0}, b_{x_0}\}$ enter only at $t = 0$, the influence tensor is initialised once per sequence,
\begin{equation}
\mathbf{M}_{W_{x_0}, ij}^{(0)} = \bigl(1 - (\bar{\xx}_i^{(0)})^2\bigr)\, s_{0,j},
\qquad
\mathbf{M}_{b_{x_0}, i}^{(0)} = 1 - (\bar{\xx}_i^{(0)})^2,
\end{equation}
and thereafter propagated forward by the multiplicative decay in Eq.~\eqref{eq:rglru_rtrl_general} with no additional immediate term, since these parameters are not reused at $t \geq 1$.

\paragraph{Recurrent parameter gradient} Given the error $\hat{\epsilon}_x^{(t)}$ returned by the inference procedure of \ref{app:rglru_inference}, the tPC-RTRL time-batched recurrent gradient accumulator is
\begin{equation}
\Delta \theta
=
\frac{1}{H}\sum_{t=1}^{T} \hat{\epsilon}_x^{(t)} \cdot \mathbf{M}_{\theta}^{(t)}
\qquad \text{for each } \theta \in \theta_{\mathrm{rec}},
\end{equation}
 and is applied at the end of the sequence as $\theta \gets \theta + \eta\,\Delta \theta$, consistent with Algorithm~\ref{alg:tpcrtrl}. The RTRL recursion, the inference dynamics, and the readout-parameter updates are all applied layer-locally to the RG-LRU cell, and no global BPTT-style reverse traversal through time is required, with spatial error propagation handled by local PC inference steps.

\paragraph{Equivalence with the two-component update}
The single-term form above is equivalent to the two-component
tPC-RTRL update of Eq.~\eqref{eq:tpc_rtrl_update} in the
predictive-rollout regime. We retain the uncollapsed two-component form in the main text for
two reasons. First, it makes the relationship to standard tPC
transparent. Standard tPC retains only the immediate term, while
tPC-RTRL augments it with the historic term, so separating these
contributions cleanly isolates the mechanism responsible for
long-range temporal credit assignment. Second, the two-component
form is what survives beyond the predictive-rollout regime. Under
inferred propagation (\ref{app:inferred-propagation}) the RTRL
recursion becomes approximate and the clean collapse into a single
term is no longer guaranteed; keeping the two components explicit makes
visible where the approximation enters (through the historic term) and which quantity remains exact regardless of regime (the immediate term).

\section{Proof that tPC-RTRL recovers BPTT in the predictive-rollout regime}
\label{app:proofs}
\label{app:bptt_equivalence}

\citet{DBLP:journals/corr/abs-2006-04182} showed that, at inference equilibrium and under the fixed-prediction assumption, the parameter updates of a PC network coincide exactly with those of an equivalent network trained by BP. We use this result to show that tPC-RTRL, as specified by the update rule stated in Eq.~\eqref{eq:tpc_rtrl_update} of the main text, recovers the gradients of a recurrent network trained by BPTT.

We state the result for the single-recurrent-layer setting, which is also the setting in which exact RTRL is most commonly analysed. The extension to multiple recurrent layers is conceptually straightforward, but would require tracking prohibitively large influence matrices. Throughout this proof, we assume the predictive-rollout regime introduced in Section~\ref{sec:method}, so that the propagated latent state is the recurrent prediction,
\begin{equation}
\bar{\xx}^{(t)} = \mu_x^{(t)} = f(\xx_L^{(t)}, \bar{\xx}^{(t-1)}; \theta_{\mathrm{rec}}),
\end{equation}
and the corresponding RTRL recursion is exact.

\paragraph{Model and objective}
Consider a recurrent latent state $\bar{\xx}^{(t)}$ together with a depth-$L$ feedforward readout network. Let the readout states be $\zz_t^{(1)}, \dots, \zz_t^{(L)}$, and define $\zz_t^{(0)} \equiv \xx^{(t)}$. The output prediction is
\begin{equation}
\bar{\yy}^{(t)} := g_L(\zz_t^{(L)}; \theta_{\mathrm{out}}^{(L)}),
\end{equation}
and the instantaneous squared-error loss is
\begin{equation}
\ell_t := \tfrac12 \|\yy^{(t)} - \bar{\yy}^{(t)}\|_2^2.
\end{equation}
Under the usual Gaussian assumptions for PC and tPC, the corresponding free energy at timestep $t$ is
\begin{align}
2\F^{(t)}
=&\;
\|\yy^{(t)} - g_L(\zz_t^{(L)}; \theta_{\mathrm{out}}^{(L)})\|_2^2 \notag\\
&+
\sum_{l=0}^{L-1}
\|\zz_t^{(l+1)} - g_l(\zz_t^{(l)}; \theta_{\mathrm{out}}^{(l)})\|_2^2 \notag\\
&+
\|\xx^{(t)} - f(\xx_L^{(t)}, \bar{\xx}^{(t-1)}; \theta_{\mathrm{rec}})\|_2^2.
\label{eq:proof_free_energy}
\end{align}
We denote the prediction errors by $\epsilon_y^{(t)} = \yy^{(t)} - g_L(\zz_t^{(L)})$, $\epsilon_l^{(t)} = \zz_t^{(l+1)} - g_l(\zz_t^{(l)})$ for $l=0,\dots,L-1$, and $\epsilon_x^{(t)} = \xx^{(t)} - \mu_x^{(t)}$.

To recover BP exactly, we adopt the fixed-prediction assumption \citep{DBLP:journals/corr/abs-2006-04182}. That is, the prediction functions and their derivatives are evaluated at the feedforward initialisations and held fixed during inference. We denote these fixed feedforward values by tildes:
\begin{align}
\tilde{\xx}^{(t)} &:= f(\xx_L^{(t)}, \bar{\xx}^{(t-1)}; \theta_{\mathrm{rec}}),
\\
\tilde{\zz}_t^{(l+1)} &:= g_l(\tilde{\zz}_t^{(l)}; \theta_{\mathrm{out}}^{(l)}),
\\
\tilde{\zz}_t^{(0)} &:= \tilde{\xx}^{(t)}.
\end{align}
In the predictive-rollout regime, the propagated state is exactly this feedforward recurrent prediction, i.e.\ $\bar{\xx}^{(t)} = \tilde{\xx}^{(t)}$.

\paragraph{Equilibrium conditions propagate prediction errors through the readout}
Inference minimises $\F^{(t)}$ with respect to the latent variables $\xx^{(t)}, \zz_t^{(1)}, \dots, \zz_t^{(L)}$. Setting $\partial \F^{(t)} / \partial \zz_t^{(L)} = 0$ gives
\begin{equation}
\epsilon_{L-1}^{(t)}
=
\epsilon_y^{(t)}\,\frac{\partial g_L(\tilde{\zz}_t^{(L)}; \theta_{\mathrm{out}}^{(L)})}{\partial \tilde{\zz}_t^{(L)}}.
\end{equation}
The same argument applied to each intermediate readout state gives the backward-propagating recursion
\begin{equation}
\epsilon_{l-1}^{(t)}
=
\epsilon_l^{(t)}\,\frac{\partial g_l(\tilde{\zz}_t^{(l)}; \theta_{\mathrm{out}}^{(l)})}{\partial \tilde{\zz}_t^{(l)}},
\qquad l = 1,\dots,L-1.
\label{eq:error_recursion}
\end{equation}
Finally, setting $\partial \F^{(t)} / \partial \xx^{(t)} = 0$ (noting $\zz_t^{(0)} \equiv \xx^{(t)}$) yields
\begin{equation}
\epsilon_x^{(t)}
=
\epsilon_0^{(t)}\,\frac{\partial g_0(\tilde{\xx}^{(t)}; \theta_{\mathrm{out}}^{(0)})}{\partial \tilde{\xx}^{(t)}}.
\label{eq:eps_x_eq}
\end{equation}
Unrolling Eq.~\eqref{eq:error_recursion} from $l=L$ down to $l=0$ and substituting into Eq.~\eqref{eq:eps_x_eq} gives
\begin{equation}
\epsilon_x^{(t)}
=
\epsilon_y^{(t)}
\prod_{l=0}^{L}
\frac{\partial g_l(\tilde{\zz}_t^{(l)}; \theta_{\mathrm{out}}^{(l)})}{\partial \tilde{\zz}_t^{(l)}}
=
\epsilon_y^{(t)}\,\frac{\partial \bar{\yy}^{(t)}}{\partial \tilde{\xx}^{(t)}},
\label{eq:eps_x_closed_form}
\end{equation}
where the last equality is the chain rule through the feedforward readout.

\paragraph{Spatial PC recovers BP at the recurrent state}
The instantaneous loss satisfies
\begin{equation}
\frac{\partial \ell_t}{\partial \tilde{\xx}^{(t)}}
=
-(\yy^{(t)} - \bar{\yy}^{(t)})\,\frac{\partial \bar{\yy}^{(t)}}{\partial \tilde{\xx}^{(t)}}
=
-\epsilon_y^{(t)}\,\frac{\partial \bar{\yy}^{(t)}}{\partial \tilde{\xx}^{(t)}}.
\end{equation}
Combining with Eq.~\eqref{eq:eps_x_closed_form} gives
\begin{equation}
\epsilon_x^{(t)}
=
-\frac{\partial \ell_t}{\partial \tilde{\xx}^{(t)}},
\label{eq:spatial_pc_eq_bp}
\end{equation}
which tells us that the spatial BP error is equal to the PC local error at this inference equilibrium. The same recursive argument also gives the standard spatial-PC result for the readout weights \cite{DBLP:journals/corr/abs-2006-04182},
\begin{equation}
\frac{\partial \F^{(t)}}{\partial \theta_{\mathrm{out}}^{(l)}}
=
\frac{\partial \ell_t}{\partial \theta_{\mathrm{out}}^{(l)}},
\qquad l = 0,\dots,L,
\end{equation}
so that PC recovers BP for all parameters in the readout path.

\paragraph{tPC-RTRL update equals the BPTT gradient at each timestep}
We now consider the recurrent parameters $\theta_{\mathrm{rec}}$. The tPC-RTRL update at time $t$, as given by Eq.~\eqref{eq:tpc_rtrl_update} of the main text, is the sum of an immediate and a historic term,
\begin{equation}
\frac{\partial \F^{(t)}}{\partial \theta_{\mathrm{rec}}}
=
\underbrace{\frac{\partial \F^{(t)}}{\partial \theta_{\mathrm{rec}}^{(t)}}}_{\text{immediate}}
+
\underbrace{\frac{\partial \F^{(t)}}{\partial \bar{\xx}^{(t-1)}}\mathbf{M}^{(t-1)}}_{\text{historic}}.
\label{eq:update_step}
\end{equation}
The free energy depends on $\theta_{\mathrm{rec}}^{(t)}$ and on $\bar{\xx}^{(t-1)}$ only through the recurrent prediction $\mu_x^{(t)}$ in the term $\|\xx^{(t)} - \mu_x^{(t)}\|^2$. Differentiating gives
\begin{align}
\frac{\partial \F^{(t)}}{\partial \theta_{\mathrm{rec}}^{(t)}}
&=
-\epsilon_x^{(t)}\,\frac{\partial \mu_x^{(t)}}{\partial \theta_{\mathrm{rec}}^{(t)}},
\label{eq:pc_Wx_grad}
\\
\frac{\partial \F^{(t)}}{\partial \bar{\xx}^{(t-1)}}
&=
-\epsilon_x^{(t)}\,\frac{\partial \mu_x^{(t)}}{\partial \bar{\xx}^{(t-1)}}.
\label{eq:pc_xbar_grad}
\end{align}
In the predictive-rollout regime, $\bar{\xx}^{(t)} = \mu_x^{(t)}$, so the RTRL recursion for the influence matrix is exact:
\begin{align}
\mathbf{M}^{(t)}
&=
\frac{\partial \bar{\xx}^{(t)}}{\partial \theta_{\mathrm{rec}}^{(t)}}
+
\frac{\partial \bar{\xx}^{(t)}}{\partial \bar{\xx}^{(t-1)}}\mathbf{M}^{(t-1)}\notag\\
&=
\frac{\partial \mu_x^{(t)}}{\partial \theta_{\mathrm{rec}}^{(t)}}
+
\frac{\partial \mu_x^{(t)}}{\partial \bar{\xx}^{(t-1)}}\mathbf{M}^{(t-1)}.
\label{eq:rtrl_recursion_in_proof}
\end{align}
Substituting Eqs.~\eqref{eq:pc_Wx_grad}, \eqref{eq:pc_xbar_grad}, and \eqref{eq:rtrl_recursion_in_proof} into Eq.~\eqref{eq:update_step} gives
\begin{align}
\frac{\partial \F^{(t)}}{\partial \theta_{\mathrm{rec}}}
&=
-\epsilon_x^{(t)}\left[
\frac{\partial \mu_x^{(t)}}{\partial \theta_{\mathrm{rec}}^{(t)}}
+
\frac{\partial \mu_x^{(t)}}{\partial \bar{\xx}^{(t-1)}}\mathbf{M}^{(t-1)}
\right] \notag\\
&=
-\epsilon_x^{(t)}\,\mathbf{M}^{(t)}.
\label{eq:update_collapsed}
\end{align}
Applying Eq.~\eqref{eq:spatial_pc_eq_bp} then gives
\begin{equation}
\frac{\partial \F^{(t)}}{\partial \theta_{\mathrm{rec}}}
=
\frac{\partial \ell_t}{\partial \tilde{\xx}^{(t)}}\,\mathbf{M}^{(t)}.
\end{equation}
In the predictive-rollout regime, $\tilde{\xx}^{(t)} = \bar{\xx}^{(t)}$, and the instantaneous loss $\ell_t$ depends on $\theta_{\mathrm{rec}}$ only through the recurrent state at time $t$. Hence by the chain rule,
\begin{equation}
\frac{\partial \ell_t}{\partial \tilde{\xx}^{(t)}}\,\mathbf{M}^{(t)}
=
\frac{\partial \ell_t}{\partial \bar{\xx}^{(t)}}\,\frac{\partial \bar{\xx}^{(t)}}{\partial \theta_{\mathrm{rec}}}
=
\frac{\partial \ell_t}{\partial \theta_{\mathrm{rec}}},
\end{equation}
and therefore
\begin{equation}
\frac{\partial \F^{(t)}}{\partial \theta_{\mathrm{rec}}^{(t)}}
+
\frac{\partial \F^{(t)}}{\partial \bar{\xx}^{(t-1)}}\mathbf{M}^{(t-1)}
=
\frac{\partial \ell_t}{\partial \theta_{\mathrm{rec}}}.
\label{eq:step_equivalence}
\end{equation}

\paragraph{Summing over time}
Summing Eq.~\eqref{eq:step_equivalence} over $t = 1, \dots, T$ gives
\begin{align}
\sum_{t=1}^{T}
\left[
\frac{\partial \F^{(t)}}{\partial \theta_{\mathrm{rec}}^{(t)}}
+
\frac{\partial \F^{(t)}}{\partial \bar{\xx}^{(t-1)}}\mathbf{M}^{(t-1)}
\right]
&=
\sum_{t=1}^{T}
\frac{\partial \ell_t}{\partial \theta_{\mathrm{rec}}}\notag\\
&=
\frac{\partial}{\partial \theta_{\mathrm{rec}}}
\sum_{t=1}^{T}\ell_t,
\end{align}
which is Eq.~\eqref{eq:prop_bptt_match} of Proposition~\ref{prop:tpcrtrl_bptt}.

Hence, under the fixed-prediction assumption and in the predictive-rollout regime, the two-term tPC-RTRL update supplies the usual spatial backpropagated error signal through the readout network via the equilibrium prediction error $\epsilon_x^{(t)}$, whilst the RTRL recursion composes the immediate and historic terms into the temporal credit-assignment signal through the recurrent latent trajectory. Their sum over time recovers the BPTT gradient.

\section{Cosine similarity between tPC-RTRL and BPTT gradients}
\label{app:cosine_similarity}

\begin{figure*}[t]
    \centering
    \includegraphics[width=0.75\linewidth]{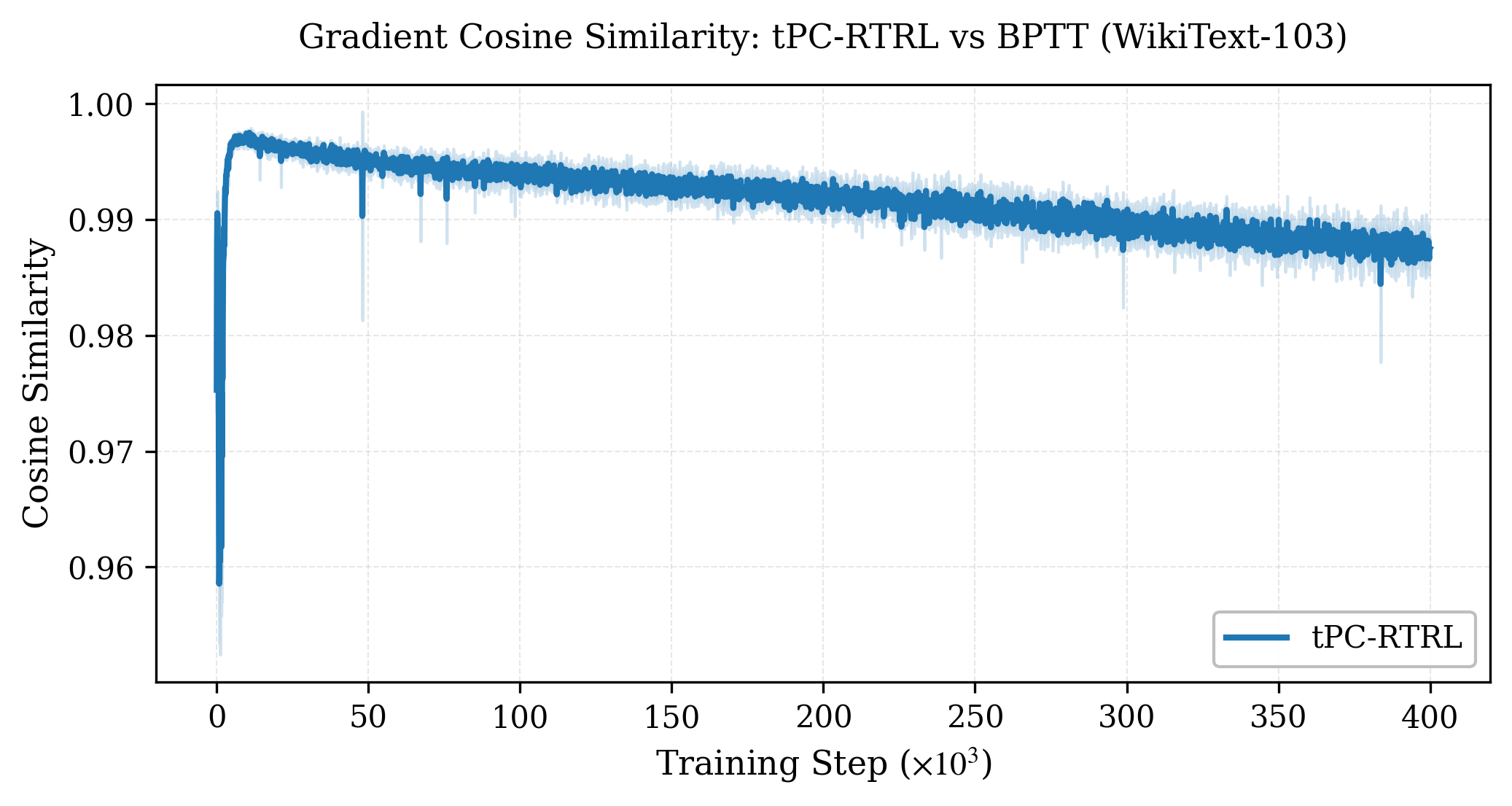}
    \caption{
        Cosine similarity between the tPC-RTRL and BPTT parameter gradient vectors on WikiText-103 (Section \ref{sec:wikitext}), as a function of training step. Both gradients are computed on the same minibatch using the current model parameters. The line shows the mean across 5 seeds; shaded region indicates one standard deviation.
    }
    \label{fig:wikitext_cosine_similarity}
\end{figure*}

The proof in \ref{app:bptt_equivalence} establishes that tPC-RTRL coincides with BPTT exactly under the fixed-prediction assumption and converged inference. As discussed in Sections~\ref{sec:bptt_relation} and~\ref{sec:near_equivalence}, the experiments in the main text do not enforce this regime, but instead, for each task, implement iterative free energy minimisation at an operating point selected to maximise the cosine similarity between the tPC-RTRL and BPTT gradients on a single batch of untrained-model data. This appendix reports the cosine similarity between the two gradients over the course of training the tPC-RTRL models on WikiText-103 (Section \ref{sec:wikitext}), providing an empirical check that the selected operating point continues to produce well-aligned gradients beyond the untrained initialisation at which it was tuned.

At each logged training step, we compute the tPC-RTRL parameter gradient and the BPTT parameter gradient on the same minibatch using the current model parameters, and report the cosine similarity between the two gradient vectors. Both gradients are flattened across all parameters before the cosine is taken. Figure \ref{fig:wikitext_cosine_similarity} shows this cosine similarity throughout training. Reported values are means across the 5 training seeds used in Section~\ref{sec:wikitext}. Note that this represents the cosine similarity between the raw gradients, after the Adam scaling step the global cosine similarity pushes even closer to 1.0 as many magnitude discrepancies are absorbed into the per-parameter learning rates. 
\section{Training with inferred propagation}
\label{app:inferred-propagation}

In the predictive-rollout regime used throughout the main text, the
propagated latent state is the recurrent prediction, $\bar{\xx}^{(t)} =
\mu_x^{(t)}$, which keeps the RTRL recursion exact and aligns the training and test-time dynamics. An alternative
is the inferred-propagation regime, in which the inferred state is
propagated forward, $\bar{\xx}^{(t)} = \hat{\xx}^{(t)}$. This is natural
in settings where inference acts as an online state corrector and the
aim is to expose the recurrent dynamics to the broader distribution
of latent states reachable under inference. However, because the
transition from $\bar{\xx}^{(t-1)}$ to $\bar{\xx}^{(t)}$ is now defined
implicitly by iterative inference rather than by a closed-form
recurrence, the RTRL recursion is no longer exact, and the influence
matrix must instead be updated via the approximation in
Eq.~\eqref{eq:tpcrtrl_influence}.

\subsection{Gradient bias under inferred propagation}
\label{app:inferred-gradient-bias}

The approximation in Eq.~\eqref{eq:tpcrtrl_influence} replaces the true
implicit transition $\hat{\xx}^{(t-1)} \to \hat{\xx}^{(t)}$ with
derivatives of the recurrent prediction function evaluated at the
inferred state. The quality of this approximation depends directly
on how far the inferred state deviates from the feedforward
prediction. When $\lVert \hat{\xx}^{(t)} - \mu_x^{(t)} \rVert$ is
small, the inferred and predicted transitions are locally similar
and the approximation is close to exact. As the deviation grows,
the implicit dynamics introduced by iterative inference increasingly
diverge from the prediction-function derivatives used by the
recursion, and the influence matrix becomes an estimate of
the true sensitivity. Consequently, the recurrent parameter gradient computed from the RTRL update appears to acquire a bias that scales
with the magnitude of the inference correction.

\begin{figure*}
            \centering
            \includegraphics[width=1\linewidth]{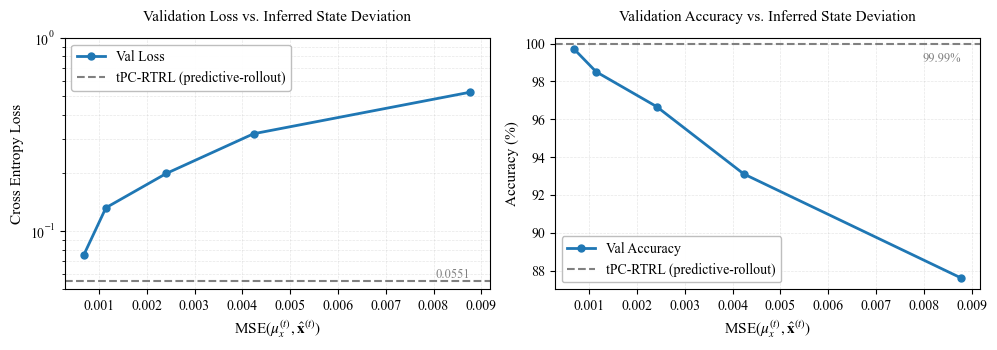}
\caption{Effect of inference deviation on copy-task performance under
inferred propagation. The number of inference iterations is swept
over $\{1, 3, 10, 20, 50\}$ at inference step size $1.0$, producing
varying mean-squared deviations $\mathrm{MSE}(\hat{x}^{(t)},
\mu_x^{(t)})$ on the horizontal axis. Left: validation cross-entropy
loss. Right: validation accuracy. For very small deviations the
inferred-propagation regime nearly matches the predictive-rollout baseline,
but performance degrades as the deviation grows, consistent
with the approximation-quality interpretation of the influence-matrix
recursion in Eq.~\eqref{eq:tpcrtrl_influence}.}
\label{fig:infprop-copy}
\end{figure*}

This dependency is visible empirically in Figure~\ref{fig:infprop-copy}
which sweeps the number of inference iterations on the copy task
whilst holding all other hyperparameters fixed\footnote{These results were obtained by evaluating the RTRL update rule at the inferred value, $\hat{\xx}^{(t)}$. Evaluating this update at the predicted state, $\mu_x^{(t)}$, led to very similar results with the same overall trend.}. Larger numbers of
inference iterations produce larger deviations between
the inferred state and the feedforward prediction. As this deviation
increases, validation loss rises and validation accuracy falls. For
very small deviations the inferred- and predictive-rollout regimes
are very similar. Beyond a modest deviation threshold,
performance degrades, and the approximation-induced bias
begins to dominate the learning signal. However, performance still remains higher than that of standard tPC, suggesting that although not perfect, the RTRL term still appears to provide useful temporal-credit information.

\subsection{Inferred propagation on the nanodrone benchmark}
\label{app:inferred-nanodrone}

\begin{figure}
    \centering
    \includegraphics[width=1\linewidth]{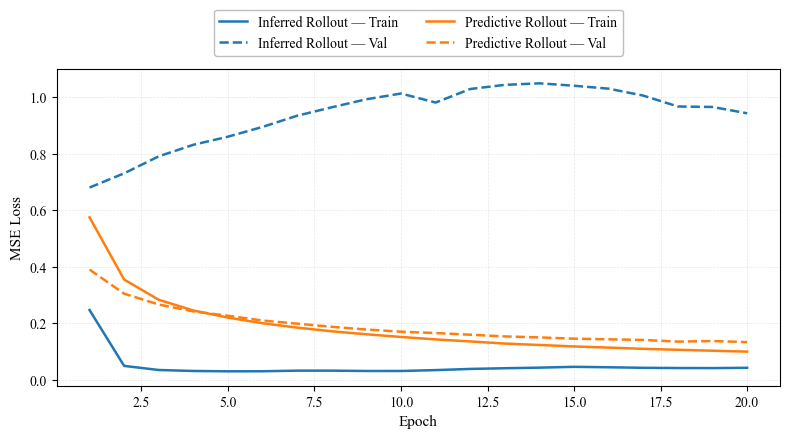}
\caption{Training and validation MSE loss on the nanodrone benchmark
under inferred propagation (blue) and predictive-rollout (orange).
Both models share architecture and optimiser settings; inferred propagation
uses $10$ iterations with step size $1.0$ and momentum $0.9$, predictive-rollout uses the setup described in Section \ref{sec:nanodrone_setup}.
Inferred propagation achieves substantially lower training loss but
markedly higher validation loss. Validation is evaluated in pure
open-loop rollout without iterative inference, so the gap reflects
a train/test mismatch where inference-based correction during
training relieves the recurrent dynamics of the need to produce
accurate autonomous rollouts.}
\label{fig:infprop-nanodrone}
\end{figure}

Figure~\ref{fig:infprop-nanodrone} applies the inferred-propagation
regime to the nanodrone benchmark of Section~\ref{sec:nanodrone}. Under
inferred propagation, the training loss drops rapidly and settles
well below the predictive-rollout baseline, yet the validation loss
remains substantially higher and shows no corresponding decrease
throughout training. Validation is evaluated in pure open-loop
rollout without iterative inference, so the validation loss is a
direct measure of the autonomous rollout quality of the learnt
recurrent dynamics.

We hypothesise that under inferred propagation the latent trajectory followed during training is
continually repaired by inference. The recurrent dynamics are
therefore never required, during training, to produce accurate
predictions from their own autonomous rollouts, and there is
correspondingly little optimisation pressure on them to do so. The
model is effectively graded on a corrected trajectory that will not
be available at test-time, which is related to the issues flagged in Section~\ref{sec:three_states}.
This means that the training loss is optimistic
because the model is graded against corrected states, and open-loop
rollout performance is poor because the training regime does not incentivise its improvement.

These observations start to help motivate the inference-aware training schedule
discussed in Section~\ref{sec:beyond_predictive}. A gradual
transition from predictive to inferred propagation during training
may allow the model to first establish stable rollout dynamics
under the exact RTRL recursion, before being exposed to the richer
distribution of latent states reachable under inference. Characterising how tPC-RTRL may perform in this setting is a concrete direction for future work.

\section{Experimental details}
\label{app:exp_details}

\subsection{Learning algorithms}
\label{app:exp_algorithms}

Across all experiments we compare four learning rules that share the
same forward model and differ only in how they assign credit to the
recurrent parameters.

\paragraph{BPTT} Standard backpropagation-through-time, with gradients
propagated through the full recurrent trajectory.

\paragraph{Spatial BP} The recurrent hidden state is detached at every
timestep, so only one-step temporal credit assignment is available.
Non-recurrent parameters are trained identically to BPTT.

\paragraph{tPC} Standard Temporal Predictive Coding in the
predictive-rollout regime. At each timestep the model produces a
recurrent prediction $\mu_x^{(t)}$ from the previous propagated state,
iterative inference minimises the instantaneous free energy
$\F^{(t)}$ to produce the inferred state $\hat{\xx}^{(t)}$, and
parameters are updated from the corresponding local errors. Recurrent
dependencies between timesteps are detached, giving the one-step
recurrent update of Section~\ref{sec:method}.

\paragraph{tPC-RTRL} Identical to tPC in its forward rollout and
inference procedure, but the recurrent parameter update is augmented
with the RTRL influence-matrix term of
Eq.~\eqref{eq:tpc_rtrl_update}, computed exactly via the recursion in
Eq.~\eqref{eq:tpcrtrl_exact}. Non-recurrent parameters are updated as
in tPC.

For the two PC-based methods, inference is performed as iterative
gradient descent on the free energy with a fixed step size and
optional momentum; the number of inference iterations and step size
are task-specific and given below. Within each seed all four methods
are initialised from the same random parameters, so any difference in
performance is attributable only to the learning rule. 

\subsection{Copy task}
\label{app:copy_task_setup}

\paragraph{Task} The vocabulary has size 10, with tokens 1--9 sampled
uniformly and 0 reserved as padding. Each example consists of a
length-30 source sequence and a target shifted by 10 steps, giving a
total sequence length of 40. The model is trained with per-timestep
cross-entropy to reproduce the delayed sequence, and inputs are
represented as one-hot vectors.

\paragraph{Architecture} All four methods share a single-layer
RNN with $\tanh$ recurrence and a linear readout and hidden size 128.

\paragraph{Training} Models are trained with Adam at learning rate
$10^{-3}$ and batch size 16. One training epoch consists of 16
minibatches, after which the model is evaluated on a fixed validation
set of 200 randomly generated samples. The training minibatches for a
given seed and epoch are materialised once and reused across all four
methods, as is the validation set. For the PC-based methods we use a
single inference iteration with step size $1.0$.

\subsection{WikiText-103}
\label{app:exp_details_wikitext}

\paragraph{Dataset} We use the raw WikiText-103 corpus
\cite{merity2016pointersentinelmixturemodels}, encoded as UTF-8 bytes
over a vocabulary of size 256. Training windows of length 257 are
sampled with replacement from the training split; the first 256 bytes
form the input and the final 256 bytes the next-byte target.
Validation and test windows are constructed deterministically by
partitioning the corresponding splits into contiguous non-overlapping
windows of the same length.

\paragraph{Architecture} The model consists of a byte embedding of
dimension $I = 512$, a single-layer RG-LRU with recurrent state size
$H = 512$, a ReLU readout of width $R = 1024$, and a linear output
head over the $C = 256$ byte vocabulary. Since $I = H$, the input
projection of \ref{app:rglru} reduces to the identity
($W_{\mathrm{in}} = \mathbb{I}$, $b_{\mathrm{in}} = 0$) and drops out
as a learnable parameter. The model has $1{,}313{,}536$ trainable
parameters in total, of which $525{,}824$ are recurrent parameters
(the RG-LRU parameters $\theta_{\mathrm{rec}} = \{\Lambda, W_a, b_a,
W_z, b_z\}$) requiring temporal credit assignment.

\paragraph{Training} Models are trained for 400{,}000 steps with batch
size 16 using Adam ($\beta_1 = 0.9$, $\beta_2 = 0.999$), learning
rate $10^{-3}$, and gradient-norm clipping at 1.0. A cosine
learning-rate schedule with 2000 warmup steps and a minimum ratio of
$0.1$ is used. For all methods, the embedding layer is
loaded from a pretrained checkpoint and kept frozen, isolating the
algorithm comparison to the recurrent and readout components. The embedding layer is trained using a single BPTT run from a different seed, the results of this run are not reported in this paper, so do not contribute towards the BPTT results of this task. For
the PC-based methods, inference uses 2 iterations with step size
$1.0$ and momentum $0.9$.

\paragraph{Evaluation} Models are evaluated every 5000 steps on
validation and test sets. Performance is reported in bits-per-character,
\begin{equation}
\mathrm{BPC} = \frac{\mathcal{L}}{\log 2},
\end{equation}
where $\mathcal{L}$ is the mean cross-entropy in nats. For each seed
the checkpoint with the best validation BPC is used to report final
test performance.

\subsection{Translation}
\label{app:exp_details_translation}

\paragraph{Dataset} We use an English--French subset of CCMatrix
\cite{schwenk2020ccmatrixminingbillionshighquality}, pre-tokenised and
stored as train/validation/test splits of sizes
$500{,}000 / 50{,}000 / 50{,}000$. A BPE tokeniser
\cite{sennrich2016neural} with vocabulary size 10{,}000 is trained
jointly on the source and target sides of the training split, and the
model uses explicit \texttt{[PAD]}, \texttt{[UNK]}, and \texttt{[EOS]}
special tokens. Within each minibatch, source--target pairs are padded
to the maximum sequence length in the batch. The data is presented to the model as concatenated source--target sequences, with an \texttt{[EOS]} token marking the boundary between the source prefix and decoder-side target tokens.

\paragraph{Architecture} The architecture mirrors that of the
WikiText-103 model at higher width. It consists of a BPE embedding of
dimension $I = 1024$, a single-layer RG-LRU with recurrent state size
$H = 1024$, a ReLU readout of width $R = 1024$, and a linear output
head over the $C = 10{,}000$ token vocabulary. As in the WikiText-103
model, $I = H$ so the input projection collapses to the identity and
drops out of $\theta_{\mathrm{rec}}$. The model has $13{,}399{,}824$
trainable parameters, of which $2{,}100{,}224$ are recurrent.

\paragraph{Training} Models are trained for 20 epochs with batch size
128 using AdamW ($\beta_1 = 0.9$, $\beta_2 = 0.999$,
$\epsilon = 10^{-8}$, weight decay $0.01$) at learning rate
$10^{-3}$, a \texttt{ReduceLROnPlateau} schedule, and gradient
clipping at norm 2.0. The training objective is token-level
cross-entropy with label smoothing $0.05$ and padding masked out. For
all methods, the embedding layer is loaded from a pretrained
checkpoint and kept frozen. The embedding layer is trained using a single BPTT run from a different seed, the results of this run are not reported in this paper, so do not contribute towards the BPTT results of this task. For the PC-based methods, inference uses
2 iterations with step size $1.0$ and momentum $0.9$.

\paragraph{Scheduled input dropout} To reduce reliance on
teacher-forcing, we apply scheduled input dropout. Let the
first \texttt{[EOS]} token mark the boundary between source and
decoder-side tokens in the concatenated input sequence. From epoch 5
onward, decoder-side input tokens are replaced with \texttt{[UNK]}
with a probability that increases linearly over the remainder of
training, up to a maximum of $0.3$. Source-side tokens are not
corrupted.

\paragraph{Evaluation} The final checkpoint is evaluated on the test
split. Test BLEU is computed by decoding each target sequence from
the source prefix terminated by \texttt{[EOS]}, using beam search
with beam size 12, length-penalty exponent $\alpha = 1.2$, repetition
penalty $1.0$, and a maximum decoded length of 128 new tokens.
Teacher-forced perplexity is computed with padding masked out.

\subsection{Nanodrone}
\label{app:exp_details_nanodrone}

\paragraph{Dataset and preprocessing} We use the data pipeline
released with the Nonlinear System Identification Nanodrone Benchmark
\cite{Busetto_2026}, which samples synchronised onboard motor
commands and motion-capture ground truth at 100\,Hz
($\Delta t = 0.01$\,s). The training split combines the
\texttt{random}, \texttt{square}, and \texttt{chirp} trajectory
families (runs 1--3), with run 4 of the same families used for
validation. The \texttt{melon} trajectory (runs 1--3) is held out
entirely as the test set. Trajectories are cut into rolling windows
of length $T = 200$ timesteps.

The prediction target is the 9-dimensional body-frame state
$s_t = [v^b_t, \phi_t, \omega_t]$ defined in
Section~\ref{sec:nanodrone_setup}: body-frame linear velocity,
orientation as a rotation vector, and body-frame angular rate.
Targets are produced by the authors' preprocessing code as supplied
with the benchmark.

\paragraph{Architecture} The model consists of three components. A
$\tanh$ state-initialisation head maps the 9-dimensional ground-truth
drone state to the initial recurrent state $\bar{\xx}^{(0)} \in
\mathbb{R}^{H}$ via Eq.~\eqref{eq:rglru_state_init}, with $W_{x_0}
\in \mathbb{R}^{H \times 9}$ and $b_{x_0} \in \mathbb{R}^{H}$. The
RG-LRU core has recurrent state size $H = 128$ and input dimension
$I = 4$, so the learnt input projection
$\{W_{\mathrm{in}} \in \mathbb{R}^{H \times I}, b_{\mathrm{in}} \in
\mathbb{R}^{H}\}$ of \ref{app:rglru} is active and trained jointly
with $\theta_{\mathrm{rec}} = \{\Lambda, W_a, b_a, W_z, b_z\}$. The ReLU
readout has width $R = 128$ and a linear output head maps to the 9-dim body-frame state. The initialisation head is needed as at the start of the sequence, the drone has some non-zero state that must be mapped into the recurrent state to produce accurate predictions. The overall model has $21{,}001$ trainable parameters.

\paragraph{Training} Models are trained for 50 epochs with batch
size 256 using Adam at learning rate $10^{-3}$ and no weight decay. A
\texttt{ReduceLROnPlateau} schedule is applied on validation loss
with reduction factor $0.5$, patience of 10 epochs, and minimum
learning rate $10^{-7}$. The training objective is the mean-squared
error between the predicted and ground-truth body-frame state,
averaged over the batch and over all timesteps in the
window. For the PC-based methods, inference uses
3 iterations with step size $1.0$ and momentum $0.9$, with
free energy terms at the recurrent state, the readout projection,
and the output all weighted equally. Checkpoints are selected by
validation loss.

\paragraph{Evaluation} The selected checkpoint is evaluated by
rolling the model out across the full 200-step \texttt{melon} windows
in open loop, initialising the recurrent state from the ground-truth
state at the start of each window. Predicted body-frame velocities
are rotated back into the world frame using the predicted rotation
vector, and world-frame position is recovered by Euler integration from the ground-truth initial position.

\subsection{Filtering inference iterations}

In the nanodrone filtering experiments, test-time correction is performed by running iterative PC inference on the recurrent state whenever an observation is available. Each correction uses 100 inference iterations and an inference learning rate of 1.0. The recurrent prediction at the current timestep is used as the initial latent state and as a prior, while the observed 9-dimensional body-frame state provides the likelihood term. After inference, the corrected latent state is propagated forward, so filtering uses inferred-state propagation at deployment time even though the model was trained in the predictive-rollout regime.

\end{document}